\documentclass[10pt,journal,compsoc]{IEEEtran}
\ifCLASSOPTIONcompsoc
  \usepackage[nocompress]{cite}
\else
  \usepackage{cite}
\fi
\ifCLASSINFOpdf
\else
\fi
\usepackage{amsmath}
\usepackage{graphicx}
\usepackage{soul}
\usepackage{physics}
\usepackage{graphicx}
\usepackage{amsmath}
\usepackage{tikz}
\usepackage{multirow}
\usepackage{array}
\usepackage{pgflibraryarrows}
\usepackage{pgflibrarysnakes}
\usepackage{booktabs}
\usepackage{url}
\usepackage{subfig}
\usepackage{amsfonts}
\usepackage{amssymb}
\usepackage{mathdesign}
\usepackage{color, xcolor}
\usepackage{pifont}
\usepackage[switch]{lineno}
\usepackage{ulem}
\begin{document}
%
\title{A Quantum Probability Driven Framework for Joint Multi-Modal Sarcasm, Sentiment and Emotion Analysis}
%
%
%

  
\author{Yaochen Liu,
  Yazhou Zhang,
  Dawei Song
\thanks{Manuscript received March 3, 2022; revised November 3, 2022 and April 19, 2023. This work is supported by Natural Science Foundation of Beijing (grant No. 4222036), National Science Foundation of China under grant No. 62006212, Foundation of Key Laboratory of Dependable Service Computing in Cyber-Physical-Society (Ministry of Education), Chongqing University(PJ.No:CPSDSC202103)the fund of State Key Lab. for Novel Software Technology in Nanjing University under grant No. KFKT2021B41, the Industrial Science and Technology Research Project of Henan Province under Grants 222102210031, 222300420582, the Doctoral Scientific Research Foundation of Zhengzhou University of Light Industry (grant No 2020BSJJ030, 2020BSJJ031).(\textit{Corresponding authors: Dawei Song and Yazhou Zhang)}}
\thanks{Y. Liu and D. Song are with the School of Computer Science and Technology, Beijing Institute of Technology, Beijing 100081, China, and D. Song is also with the School of Computing and Communications, The Open University, Walton Hall, United Kingdom (email: dawei.song2010@gmail.com).}
\thanks{Y. Zhang is with School of Nursing, The Hong Kong Polytechnic University. He is also with Key Laboratory of Dependable Service Computing in Cyber Physical Society (Chongqing University), Ministry of Education; Artificial Intelligence Laboratory, China Mobile Communication Group Tianjin Co., Ltd.; State Key Lab. for Novel Software Technology, Nanjing University, Nanjing, China; Zhengzhou University of Light Industry (email: yazzhang@polyu.edu.hk).}
\thanks{Yaochen Liu and Yazhou Zhang contribute equally and share the co-first authorship.}}

%
%

\markboth{IEEE TRANSACTIONS ON AFFECTIVE COMPUTING, XX-XX}%
{Liu \MakeLowercase{\textit{et al.}}: A Quantum Probability Driven Framework for Joint Multi-Modal Sarcasm, Sentiment and Emotion Analysis }
%



\IEEEtitleabstractindextext{%
\begin{abstract}
Sarcasm, sentiment, and emotion are three typical kinds of spontaneous affective responses of humans to external events and they are tightly intertwined with each other. Such events may be expressed in multiple modalities (e.g., linguistic, visual and acoustic), e.g., multi-modal conversations. Joint analysis of humans' multi-modal sarcasm, sentiment, and emotion is an important yet challenging topic, as it is a complex cognitive process involving both cross-modality interaction and cross-affection correlation. 
From the probability theory perspective, cross-affection correlation also means that the judgments on sarcasm, sentiment, and emotion are incompatible. 
However, this exposed phenomenon cannot be sufficiently modelled by classical probability theory due to its assumption of compatibility. 
Neither do the existing approaches take it into consideration. In view of the recent success of quantum probability (QP) in modeling human cognition, particularly contextual incompatible decision making, we take the first step towards introducing QP into joint multi-modal sarcasm, sentiment, and emotion analysis. Specifically, we propose a \textbf{QU}antum probab\textbf{I}lity driven multi-modal sarcasm, s\textbf{E}ntiment and emo\textbf{T}ion analysis framework, termed QUIET.
Extensive experiments on two datasets and the results show that the effectiveness and advantages of QUIET in comparison with a wide range of the state-of-the-art baselines. We also show the great potential of QP in multi-affect analysis.  
\end{abstract}

\begin{IEEEkeywords}
Quantum Probability, Sarcasm Detection, Sentiment Analysis, Emotion Recognition, Multi-Modal Framework.
\end{IEEEkeywords}}

\maketitle
\IEEEdisplaynontitleabstractindextext

%
\IEEEpeerreviewmaketitle

\IEEEraisesectionheading{\section{Introduction}\label{sec:introduction}}

\IEEEPARstart {T}{he} ability of affect understanding and cognition is one of the main differences between human and machine. As an active research direction in AI (Artificial Intelligence), affect analysis aims to help machine infer and understand human affect, and then make an appropriate response~\cite{9512296}. Human affect is often multi-modal (e.g., language, facial expressions and acoustic behaviors) and contextual (e.g., the same utterance expresses different affects under different contexts) in nature.
 Hence, multi-modality and contextuality would provide richer clues for detecting human affect. 

As a generalized concept, human affect consists of different types of feelings, e.g., sarcasm, sentiment, emotion, etc. They are correlative and interdependent.
Sarcasm is a subtle form of metaphorical affection, where the literal meaning of the author is contrary to his/her true attitude. It often expresses criticism, anger or mock emotion. Sentiment is a long-term subjective attitude of a human based on his/her feeling towards a situation, topic or event, while emotion refers to a strong but unabiding physiological feeling such as happiness, anger and sadness.

There have been a rich body of existing approaches in multi-modal sarcasm, sentiment, and emotion analysis, most of which focus on 
multi-modal feature extraction and multi-modal fusion. For instance, Zhang et al.~\cite{zhang2018quantum} presented a quantum inspired decision fusion model for multi-modal sentiment analysis.
Hu et al. \cite{hu2021mmgcn} presented a graph neural network for conversational emotion recognition. 
The potential of analyzing sentiment, sarcasm, and emotion under a unified framework still needs the deeper exploration.

From a cognitive perspective, the analysis of sarcasm, sentiment, and emotion involves a complex cognitive phenomenon in constructing the same affect. Hence, sarcasm, sentiment, and emotion are closely intertwined with each other~\cite{chauhan2020sentiment}. For example, the sarcastic utterance ``I like work, it fascinates me. I can sit and look at it for hours.'' expresses the author's negative sentiment and a strong dislike or unhappiness of his/her job. Detecting sarcasm, sentiment, and emotion would bring benefits to each other.

From a probability theory perspective, the above phenomenon means that the judgments on sarcasm, sentiment, and emotion are incompatible, i.e., they do not share a common probability space and their joint probability cannot be determined from the marginal probabilities without considering the interference and incompatibility between these judgements. Such exposed phenomenon cannot be sufficiently modelled by classical probability theory due to its assumption of compatibility. Neither do the existing approaches take it into consideration. The recent studies have largely neglected their cognitive correlation. This raises our research question: \textit{Can we solve this cross-affection correlation and propose a multi-modal, contextual and joint multi-affect detection framework?}

To fill this gap, it is essential to jointly study multi-modal sarcasm, sentiment, and emotion from a more general cognitive framework that unifies both multi-modality, contextuality and multi-affect judgment. In recent years, quantum probability (QP), as a mathematical framework of quantum physics that proposes two assumptions of both compatibility and incompatibility, has been adopted for describing elusive human cognitive and emotional activities, where a new research community, $viz.$ quantum cognition, has been emerging\cite{bruza2015quantum
}. An increasing body of theoretical and empirical evidence has shown the effectiveness and advantages of QP in modeling various AI tasks involving human cognition, 
e.g., semantic analysis,
 question answering
and sentiment classification.
 For instance,  quantum language model (QLM)~\cite{sordoni2013modeling} 
represented 
user's information needs and documents as density matrices (DMs) in a quantum probabilistic space. 
Wang and Li et al.~\cite{li2019cnm} defined a complex semantic Hilbert space to capture the ``quantumness'' in the cognitive aspect of human language.

In this paper, we theoretically justify the use of QP in the multi-modal sarcasm, sentiment and emotion analysis task. Then, we propose a \textbf{QU}antum probab\textbf{I}lity driven multi-modal sarcasm, s\textbf{E}ntiment and emo\textbf{T}ion analysis framework, termed QUIET.
Specially, it consists of a complex-valued multi-modal encoder, a quantum composition layer, a quantum interference-like inter-modal fusion layer and a quantum measurement layer. It is formulated and applied to conversational multi-modal multi-affection detection. 
First, each multi-modal (e.g., textual, visual and acoustic) utterance is encoded as a quantum superposition of a set of basis terms, represented as a complex-valued vector. 
Second, three complex-valued vectors are fed into the quantum composition layer to learn their contextual representations. 
Third, all contextual representations are forwarded to the quantum interference-like fusion layer for producing a fused multi-modal representation. 
Finally, quantum incompatible measurements are performed on the multi-modal representation to yield the probabilistic outcomes of sarcasm, sentiment, and emotion recognition. 

We verify the effectiveness of the QUIET framework on two benchmark datasets, i.e., MUStARD and MELD. Extensive empirical results demonstrate the potential of using QP, with the QUIET framework outperforming the existing state-of-the-art approaches by large margins. The major innovations of the work are summarized as follows.

\begin{itemize}
\item We propose a quantum probability driven multi-task learning framework for joint multi-modal sarcasm, sentiment and emotion analysis, aiming to address the challenges of multi-modal affect understanding.

\item We propose a multi-modal complex-valued representation approach by leveraging the concept of quantum superposition.

\item We design a quantum-like fusion network to effectively model both intra-modality contextuality and inter-modality incongruity.

\item We present the theoretical advantages of our QUIET model, and further empirically show its effectiveness on two benchmark datasets.
\end{itemize}

The rest of this paper is organized as follows. Section 2 briefly depicts the related work. Section 3 presents the preliminaries of QP. Section 4 provides a detailed theoretical interpretation for the advantages of using QP in the multi-modal sarcasm, sentiment and emotion analysis task. Section 5 describes the proposed QUIET model step by step. In Section 6, we report the empirical experiments and conduct a detailed analysis. We conclude the paper and discuss  future research directions in Section 7.

\section{Related Work}
In this section, we review the related works on sentiment analysis, sarcasm detection and emotion recognition.

\subsection{Sentiment Analysis}
Sentiment analysis refers to the study, analysis and identification of the subjective polarity carried in user generated contents. Now, deep learning based approaches have been widely proposed. For instance,
to solve the problem of sentiment reversal,
Wang et al.~\cite{8700266} proposed an iterative algorithm called SentiDiff for Twitter sentiment analysis.
Under the inspiration that linguistic hints can serve as reliable polarity indicators, Wang et al.~\cite{8869780} proposed a joint framework, termed SenHint, which could 
integrate the representation vector and implications of linguistic hints into a unified model.
Zhang et al.~\cite{9415156} emphasized on the need of incorporating the correlation among multiple domains, and proposed an efficient adaptive transfer network (EATN) for aspect-based sentiment analysis.
Inspired by quantum theory (in short QT), Zhang et al.~\cite{zhang2018unsupervised} first used density matrix to represent textual word and designed quantum relative entropy to detect sentiment via an unsupervised manner. Their model did not consider the contextual information nor the multi-modal fusion.

\subsection{Sarcasm Detection}
Sarcasm detection is a relatively less explored task, as sarcasm often completely flips the sentiment polarity of a sentence. Nowadays, the mainstream approaches can be divided into two categories, which are traditional machine learning based methods that take the feature engineering work apart from the classification, and deep neural networks based methods that unified the feature engineering and classification task.

Ashwin et al.~\cite{Ashwin2015Behavioral} constructed a behavioral modeling framework using the behavioral and psychological features. They used the user's historical tweets as behavioral intrinsic traits, and evaluated the framework on sarcastic tweets. 
Aditya et al.~\cite{joshietal2016harnessing} targeted at using sequential features of a scene to predict sarcasm for each utterance in conversations. The proposed sequence labeling algorithms (SVM-HMM and SEARN) outperformed three traditional classification-based algorithms.

As deep learning based architectures cast off the fetters of feature engineering, they usually achieve better performance.
Leveraging the multi-modal sentiment and emotion information, Chauhan et al.~\cite{chauhan2020sentiment} used a segment-wise inter-modal attention based framework for sarcasm detection.
Zhang et al.~\cite{zhang2022stance} first used the stance information to detect sarcasm, and proposed a new sub-task, i.e., stance based sarcasm detection.

\subsection{Emotion Recognition}
Emotion recognition is treated as a complicated and fine-grained classification task in affective computing. 
Hu et al.~\cite{hu2021mmgcn} dealt with conversational emotion recognition task via a fused graph convolutional network, which could effectively utilize dependencies and leverage speaker information.
Xie et al.~\cite{xie-etal-2021-knowledge-interactive} proposed a knowledge interactive network via the paradigm of multi-task learning, namely KI-Net. KI-Net could apply both commonsense knowledge and sentiment lexicon to enrich semantic information.
Sun and Yu~\cite{sun-etal-2021-discourse-aware} leveraged the discourse structures in multi-party conversation, and proposed a discourse-aware graph neural network to recognize emotion. 
Focusing on the multi-label emotion detection in a multi-modal scenario, Zhang et al.~\cite{zhang-etal-2020-multi} designed a multi-modal sequence-to-set approach to model label dependency and modality dependency.

\subsection{Summary}
In summary, remarkable progress has been made in the three relevant areas, and motivated our work. However, these three areas have been studied separately. Different from the previous studies, we take sarcasm detection, sentiment analysis and emotion recognition into consideration via a multi-task learning framework. In addition, we aim to model the inter-modality interference and the correlation between sarcasm, sentiment, and emotion, from the cognitive perspective. QP has been proven to provide a generalized and unified formalism for the task. Specifically, we propose a QP inspired end-to-end multi-task learning framework for joint sarcasm, sentiment, and emotion analysis over multi-modal conversations.  

\textbf{The difference from previous QP based model.} Our work is quite different from other QP based models. We are the first to introduce quantum interference to perform three modal fusion. Moreover, we propose a new quantum incompatible measurement approach to model the cross-affect correlation. Detailed discussion has been provided in Sec.6.8.

\section{Quantum Probability Preliminaries}
QP offers us a mathematical and conceptual framework on capturing the intrinsically uncertain microscopic particle behaviours. Recently it has shown to be effective in modeling human fundamentally uncertain cognitive and decision making processes. In this Section, we will briefly introduce the key concepts of QP, followed by Section 4 showing how QP is suitable to model a few typical problems in human affect understanding, and thus inspires us to propose a multi-task framework.

\textbf{\textit{Quantum Superposition and Density Matrix.}} 
Quantum probability, which can be regard as a generalization of the classical probability theory. 
The mathematical base of quantum probability is established on a complex Hilbert Space, denote as $\mathcal{H}$. A quantum state vector $u$ is expressed as a ket $\ket{u}$ , its transpose is expressed as a bra $\bra{u}$. The inner product and outer product of two state vectors $\ket{u}$ and $\ket{v}$ are denoted as $\bra{u}\ket{v}$ and $|u\rangle\langle v|$. 
\textit{Quantum superposition} states that a pure quantum state can be in multiple mutually exclusive basis states simultaneously with a probability distribution until it is measured. A \textit{quantum mixture} of states gives rise to a mixed state represented by a density matrix, $\rho=\sum_i p_i |u\rangle\langle u|$, where $p_i$ denotes the probability distribution of each pure state. 

\textbf{\textit{Quantum Interference.\footnote{A detailed introduction of the double-slit experiment is given in the appendix.}}} In the double-slit experiment, a microscopic particle starts from the initial point to get to the screen through two slits at the same time or through one slit. The path difference causes a phase shift of the quantum states and produces the interference phenomena.

%
%

In the double-slit experiment, two paths interfering with each other affects the probability distribution of the particle reaching the final position on the detection screen, and forms the interference pattern. This phenomenon cannot be explained sufficiently with classical theory. We use the wave function $\varphi(x)$ to interpret this behavior. The wave function represents the probability amplitude of a particle be at a position $x$, and the square of the wave function represents the possibility. The state of the photon is in a quantum superposition of the state of path 1 and path 2, which could be formulated as: $\varphi_p (x)= \alpha \varphi_1 (x) + \beta \varphi_2 (x) $,
where $\varphi_1 (x)$  and $\varphi_2 (x)$ are the wave functions of path1 and path2. $\alpha$ and $\beta$ are complex numbers, satisfying $\alpha^2 + \beta^2 =1$. $\alpha^2$, $\beta^2$ represent the probability of the particle passing through the path1 or path2. The probability for a particle be at the state $\varphi_p$ can be calculated as: 
\begin{equation} 
\begin{aligned}
  P (x)&=|\varphi_p(x)|^2=|\alpha\varphi_1(x)+\beta\varphi_2(x)|^2\\
  \quad\quad\,\,\,&=|\alpha\varphi_1(x)|^2+|\beta\varphi_2(x)|^2\\
  &+2|\alpha\beta\varphi_1(x)\varphi_2(x)|\cos{\phi}\\
\end{aligned}
\label{eq:inter2}
\end{equation}
where $\phi$ is the interference angle. $I=2|\alpha\varphi_1(x)\beta\varphi_2(x)|\cos{\phi}$ is the interference term, which describes the interaction between two paths.

\textbf{\textit{Quantum Measurement.}} 
Measurement in the classical theory is considered to has no influence on the measured object. However, measurement in QP has an impact on the system to be measured, such as changing its state. Quantum measurement is described by a set of measurement operators, denoted as $\left \{ M_{m} \right \}$, acting on the state space of the system being measured, where $m$ represents the possible measurement outcomes. Suppose the quantum system is in a state of $\ket{u}$, then the probability to obtain the outcome $m$ after the measurement is $p\left ( m \right )= \langle u|M_{m}^{\dagger }M_{m}|u\rangle$. The Gleason's Theorem has proven the existence of a mapping function $M \left ( |u\rangle\langle u|\right )= tr\left ( \rho |u\rangle\langle u|\right )$ for any event $|u\rangle\langle u|$. Quantum measurement describes the interaction (coupling) between a quantum system and the measurement device, where the coupling system can be represented by the tensor product of two systems, e.g., $M\otimes |u\rangle$. Quantum measurement subjects to two rules: (1) for an elementary event $|u\rangle\langle u|$, $M(|u\rangle\langle u|)\in[ 0, 1 ]$. (2) for any orthogonal basis $\{ \ket{e_i} \}$, $ \sum_i^n M(|e_i\rangle\langle e_i|)=1 $. 

\section{Theoretical Justification of QP in Our Task}\label{sec:prove}

In this work, we target at dealing with three correlated tasks (sarcasm detection, sentiment analysis and emotion recognition) simultaneously. Without losing generality, we focus on utterance-level analysis of multi-modal conversation data, where each conversation consists of a sequence of multi-modal utterances. For each utterance, the determination of its sarcasm, sentiment, and emotion is inherently complex and uncertain, influenced by three key factors: its context (e.g., the historical utterances), the interaction between modalities and the correlation between the tasks. Next, we will illustrate how QP is more suitable to model such influences via theoretical justifications. We clarify that the reasons we provide such justifications are: (1) providing theoretical evidence and solid foundation; (2) letting our motivation be easy to follow. We also argue that one can also understand our proposed model even though overlooking such justifications.

\subsection{Quantum probability is more general to capture the uncertainty in human affect}
Let $z(x)=re^{i\theta}$ be a quantum complex probability amplitude of event $x$. Using the definition of quantum probability, we get the classical probability of event $x$
\begin{equation} 
	\begin{aligned}
		p(x) = |z(x)|^2 = r^2
	\end{aligned}
\end{equation}
that means
\begin{equation} 
	\begin{aligned}
		r = \sqrt{p(x)}
	\end{aligned}
\end{equation}
where $r\in\mathcal{R}$, $\theta\in(-\pi, \pi)$. Given $p(x)$, the complex probability amplitude will satisfy
\begin{equation} 
	\begin{aligned}
		z(x) = \sqrt{p(x)}  \times (\cos\theta+i\sin\theta) = re^{i\theta}
	\end{aligned}
\end{equation}

 This defines a many-to-one relationship between complex probability amplitude and probability.


\textbf{Explanation.} We have known that there is a many-to-one mapping between quantum probability amplitude and classical probability. Different quantum probability amplitude can get the same classical probability. For example, the probability of a word $w$ is 0.5, i.e., $p\left ( x=w \right )=\frac{1}{2}$, then the quantum probability amplitude may be $z\left ( x=w \right )=\frac{\sqrt{2}}{2}e^{i\frac{\pi }{4}}$ or $z\left ( x=w \right )=-\frac{\sqrt{2}}{2}e^{i\frac{3\pi }{5}}$, etc. This shows that quantum probability is more general than classic probability. The amplitude $r$ links to the probability, while the phase $\theta$ may be associated with hidden sentiment or sarcasm  orientations. An utterance thus could be represented in an amplitude-phase manner. These proofs supports our first argument that QP is advantageous in modeling the uncertainty in human language, and also answer the research question, i.e., why use quantum theory to development a multi-modal sarcasm, sentiment, and emotion model.

\subsection{Quantum interference embodies a non-linear multi-modal fusion}
Let $z_1(w_1)$ and $z_2(w_2)$ be the complex probability amplitudes of two basis words $w_1$, $w_2$ respectively \footnote{$w_1$ and $w_2$ are from different modalities, e.g., $w_1$ is a basis word from the textual modality and $w_2$ is a basis word from acoustic modality.}, where $z_1(w_1), \; z_2(w_2)\in \mathcal{H}^{l_t\times d_t}$.
Let a compound term be $c \propto (w_1,w_2)$, we obtain
\begin{equation} 
	\begin{aligned}
		z_3(c)=\alpha &z_1(w_1)+\beta z_2(w_2) \\
		s.t \quad \alpha^2 &+\beta^2=1, \\
        \alpha, \beta &\in \mathcal{C}
	\end{aligned}\label{eq:eq5}
\end{equation}
where $z_3(c)\in\mathcal{H}^{l_t\times d_t}$. Based on 
justification in \textit{Section 4.1}
, we have
\begin{equation} 
	\begin{aligned}
		p(w_1)=|\alpha|^2|z_1(w_1)|^2 &, p(w_2)=|\beta|^2|z_2(w_2)|^2 \\
		s.t\quad p(w_1),p(w_2) &\in [0,1]
	\end{aligned}\label{eq:eq6}
\end{equation}
We can derive the probability of the compound term:

\begin{equation} 
	\begin{aligned}
		p(c) &=|z_3(c)|^2=|\alpha z_1(w_1)+\beta z_2(w_2)|^2 \\
		&=(\alpha z_1(w_1)+\beta z_2(w_2))\cdot (\alpha z_1(w_1)+\beta z_2(w_2))\dagger \\
		&=\alpha z_1(w_1)\cdot(\alpha z_1(w_1))\dagger + \beta z_2(w_2)\cdot(\beta z_2(w_2))\dagger \\
		  &+\alpha z_1(w_1)\cdot(\beta z_2(w_2))\dagger +(\alpha z_1(w_1)\cdot(\beta z_2(w_2))\dagger)\dagger \\
		&=|\alpha z_1(w_1)|^2+|\beta z_2(w_2)|^2 \\
		&+2Re(\alpha z_1(w_1)\cdot(\beta z_2(w_2))\dagger) \\
		&=|\alpha z_1(w_1)|^2+|\beta z_2(w_2)|^2+2|\alpha z_1(w_1)\beta z_2(w_2)|\cos\theta \\
		&=p(w_1)+p(w_2)+2\sqrt{p(w_1)p(w_2)}\cos\theta
	\end{aligned}\label{eq:eq7}
\end{equation}

Hence, the probability of multi-modality is a non-linear combination of the two probabilities, with an interference term determined by the relative phase.

\textbf{Explanation.} The probability of the compound term is the non-linear superposition of the probabilities of the basis words, with an interference term determined by the relative phase $\theta$. This provides a higher level of abstraction. It is well known that sarcastic/sentiment/emotion expression in human language also exposes the non-linearity. For example, ``Jack leg'', which is the combination of the word ``jack'' and ``leg'', expresses an incompetent human, rather than ``jack's leg''. The linear combination of ``jack'' and ``leg'' cannot capture such abstract meaning. However, quantum interference inspired approach is able to learn the non-linear fusion. These proofs answer the research question, i.e., why use quantum interference to capture multi-modal fusion. In equation 7, $z1(w1)$ and $z2(w2)$ are complex probability amplitudes of two basis words $w1$, $w2$. Here $w1$ and $w2$ represent basis word from different modalities, e.g., $w1$ represent basis word in textual modality, and $w2$ represent basis word in acoustic modality. And to fuse all three modalities, we apply three quantum interference inspired multi-modal fusion component (t+v, t+a, v+a), after having three bi-modal representations, we concatenate them together to get tri-modal representation. 

\subsection{Quantum composition captures the contextuality between utterances}
Let $u_i$ and $u_j$ represent two adjacent utterances with the help of Dirac notation, we obtain

\begin{equation} 
	\begin{aligned}
		\ket{u_i} &= \alpha_1\ket{w_1} + \beta_1\ket{w_2} \\
		\ket{u_j} &= \alpha_2\ket{w_1} + \beta_2\ket{w_2} \\
		s.t \quad&\alpha_1^2+\beta_1^2=1, \\
		& \alpha_2^2+\beta_2^2=1, \\
		&\alpha_1, \alpha_2,\beta_1,\beta_2 \in \mathbb{C}
	\end{aligned}
\end{equation}
here $\alpha_1$, $\alpha_2$, $\beta_1$, $\beta_2$ are probability amplitudes expressed in the complex polar
form. State space of a composite system $\mathcal{H}_{u_i,u_j}$, constructed from two utterances $u_i$ and $u_j$ is written as a tensor product of individual state spaces $\ket{u_i}$ and $\ket{u_j}$:
\begin{equation} 
	\begin{aligned}
	\mathcal{H}_{u_i,u_j} &= \ket{u_i}\otimes \ket{u_j} \\
		&=(\alpha_1\ket{w_1}+\beta_1\ket{w_2})\otimes(\alpha_2\ket{w_1}+\beta_2\ket{w_2}) \\
		&=\alpha_1\ket{w_1}\otimes(\alpha_2\ket{w_1}+\beta_2\ket{w_2}) \\
		&+\beta_1\ket{w_2}\otimes(\alpha_2\ket{w_1}+\beta_2\ket{w_2}) \\
		&=\alpha_1\alpha_2\ket{w_1 w_1}+\alpha_1\beta_2\ket{w_1 w_2} \\
		&+\beta_1\alpha_2\ket{w_2 w_1}+\beta_1\beta_2\ket{w_2 w_2}
	\end{aligned}
\end{equation}
Let $\ket{w_1}=(x_1,x_2)^T$,$|w_2\rangle=(y_2,y_2)^T$, then
\begin{equation} 
	\begin{aligned}
		\mathcal{H}_{u_i,u_j} &= \alpha_1\alpha_2 
		\begin{bmatrix}
			x_1^2 & x_1x_2 \\
			x_2x_1 & x_2^2
		\end{bmatrix} 
		+\alpha_1\beta_2 
		\begin{bmatrix}
			x_1y_1 & x_1y_2 \\
			x_2y_1 & x_2y_2
		\end{bmatrix} \\
	&+\beta_1\alpha_2
		\begin{bmatrix}
			y_1x_1 & y_1x_2 \\
			y_2x_1 & y_2x_2
		\end{bmatrix}
	+\beta_1\beta_2
		\begin{bmatrix}
			y_1^2 & y_1y_2 \\
			y_2y_1 & y_2^2
		\end{bmatrix}
	\end{aligned}
\end{equation}
where $\mathcal{H}_{u_i,u_j}$ is controlled by the basis words.

\textbf{Explanation.} We observe that quantum composition treats the contextuality between utterances as the contextuality between words, which inspires us to model the contextuality by a ``global to local'' way. Our proposed approach is quite distinct from the existing context modeling approaches (e.g., adding, concatenation, etc.), which leverages the advantage of tensor to model the interaction across adjacent utterances.

\subsection{Quantum incompatible measurement describes the correlations across multi-tasks}
Let us have two observables \textit{Sen} and \textit{Sar}, represented by $M^e$ and $M^a$. Let ${|k_n\rangle}$ be a complete set of common eigenkets of the two compatible obervables $Sen$ and $Sar$, corresponding to the sets ${e_n}$ and ${a_n}$. Then,
\begin{equation} 
	\begin{aligned}
		M^eM^a\ket{k_n}&=M^ea_n\ket{k_n}=e_na_n\ket{k_n} \\
		&=a_ne_n\ket{k_n}=M^aM^e\ket{k_n}
	\end{aligned}
\end{equation}
Based on this, we obtain
\begin{equation} 
	\begin{aligned}
		(M^eM^a-M^aM^e)\ket{k_n}=0
	\end{aligned}
\end{equation}
This implies $[M^e,M^a]=0$ \footnote{The mutation rule means $[M^e,M^a]=M^eM^a-M^aM^e=0$}, which means two operators are compatible. Otherwise, if $[M^e,M^a]\neq 0$ we say two operators are incompatible. In other words, they do not satisfy the mutation rule.

\textbf{Explanation.} These proofs can be used to clarify a fact that one's sentiment judgement toward an utterance after his/her sarcasm judgement may be different from his/her sentiment judgment before sarcasm judgment. The order of judgment indeed affect the sarcasm and sentiment understanding. Quantum incompatibility help one understand the correlation across different tasks by providing a quantified metric of incompatibility measuring.

To sum up, we argue that the above-mentioned proofs provide solid foundation of our proposed model. Knowing this will deepen the understanding of 
our motivation and multi-affective joint analysis. This also gives a good answer to many readers' question: why use quantum theory to design a macro NLP model?

\section{The Proposed QUIET Model}
In this section, we detail the proposed QUIET model which leverages textual, visual and acoustic information.
\begin{figure*}[th]
	\centering
	\includegraphics[width=7.04in, height=5.12in]{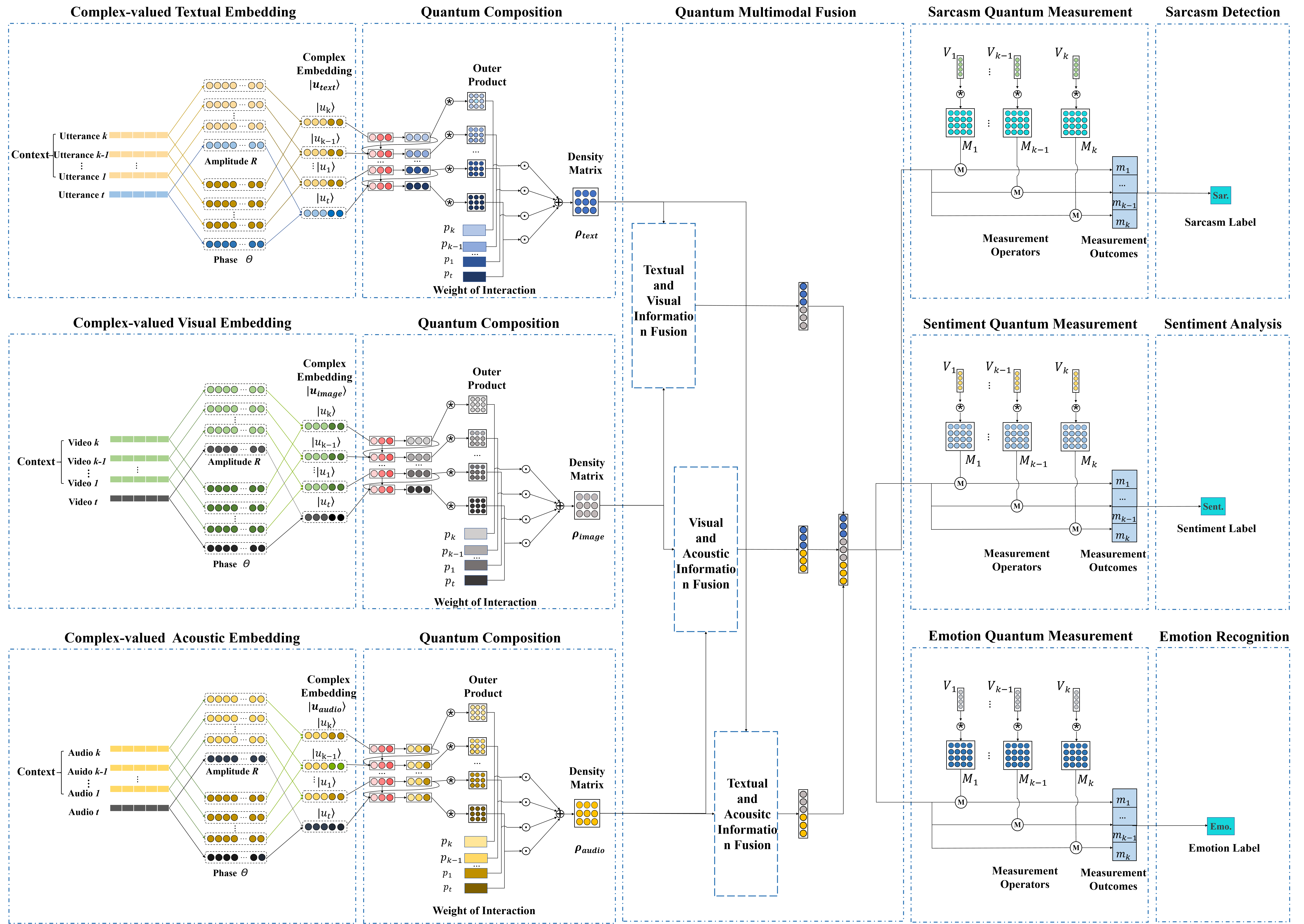}
	\caption{The architecture of the QUIET framework. $\circledast$ denotes an outer production to a vector. $\odot$ denotes point-wise multiplication.$\oplus$ refers to a element-wise addition.
	$\textcircled{M}$ 
	refers to the quantum measurement operation.}
	\label{fig:QUIETmodel}
\end{figure*}

\subsection{Task Definition}
This paper aims to detect sarcasm, sentiment, and emotion simultaneously in multi-modal conversations, via a quantum probability inspired multi-task learning framework. Assume that the dataset has $M$ samples, the $i^{th}$ sample $X^i$ is represented as $\left \{X^i = (C^i, T^i), Y^i\right \}$, where $C^i$, $T^i$, $Y^i$ respectively denote the contextual utterances, the $i^{th}$ target utterance and the sarcasm/sentiment/emotion label. Each utterance consists of three modalities, i.e., textual, visual and acoustic modalities. Suppose there are $R$ contexts for the $i^{th}$ sample, then the $f^{th}$ contextual utterance is represented as $C^i_f = (C^i_t, C^i_v, C^i_a)$, the $i^{th}$ target utterance is denoted as $T^i = (T^i_t, T^i_v, T^i_a)$, where $i \in \left[1,2,...,M\right]$, $f \in \left[1,2,...,R\right]$. The labels of $i^{th}$ target utterance are $Y^i = (y^i_{sar}, y^i_{sent}, y^i_{emo})$, describing the results for sarcasm detection, sentiment analysis and emotion recognition.

Based on the above description, the task could be formulated as:
\begin{equation}  \small
\setlength{\abovedisplayshortskip}{0.25cm} 
\setlength{\belowdisplayshortskip}{0.25cm}
\setlength{\abovedisplayskip}{0.25cm}
\setlength{\belowdisplayskip}{0cm}
\begin{aligned}
\zeta=\prod_{i}p\left ( Y^{i}|C^{i},T^{i},\Theta  \right )
\end{aligned}
\end{equation} 
where $\Theta$ represents the parameter set in the model.

\subsection{Overall Network}
The architecture of the QUIET framework is shown in Figure~\ref{fig:QUIETmodel}. It is composed of five building blocks, i.e., a complex-valued multi-modal encoder, a quantum composition layer, a quantum interference-like inter-modal fusion layer, a quantum incompatible measurement layer and a dense layer. The framework works in the following procedure. (1) The $k^{th}$ textual utterance, video clip and acoustic segment are represented by complex-valued embeddings, e.g., $\ket{T^{k}_{t}}$, $\ket{T^{k}_{v}}$ and $\ket{T^{k}_{a}}$. The technical details on initialization of these embedding vectors are provided in Section 6.1.
(2) Then, $\ket{T^{k}_{t}}$, $\ket{T^{k}_{v}}$ and $\ket{T^{k}_{a}}$ are fed into the quantum composition layer to calculate the intra-modality contextuality, where the results are encapsulated in three density matrices $\rho_{text}$, $\rho_{img}$ and $\rho_{auc}$. (3) We then fuse any two density matrices from $\rho_{text}$, $\rho_{img}$ and $\rho_{auc}$, to obtain the bi-modal representations via quantum interference. The tri-modal representation is obtained by merging them together. (4) We extract the final sarcastic, sentimental and emotional features via quantum incompatible measurement, and finally (5) we feed such features into the fully connected softmax layers to yield sarcasm, sentiment, and emotion predictions respectively. 

\subsection{Complex-valued Multi-Modal Encoder}
Motivated by Wang and Li's work~\cite{li2019cnm}, we seek inspirations from QP, and design a complex-valued multi-modal encoder. 
For text, we assume that the textual Hilbert space $\mathcal{H}_t$ is spanned by a set of orthogonal basis states $\{|w^{j}_{t}\rangle\}^{n}_{j=1}$. But different from their assumption that treats the sememes as basis states, we treat words in the textual counterpart as basis states.
In this way, the $j^{th}$ word $w^j_t$ is a basis state $|w^j_t\rangle$ in the textual Hilbert space, and is described using an one-hot encoding, which means the $j^{th}$ position is $1$ while other positions are $0$, i.e., $|w^{j}_{t}\rangle = \left ( 0, 0,...,\underset{j-1}{0},\underset{j}{1},\underset{j+1}{0},...,0 \right )^{T}$. 
Then, the $k^{th}$ target utterance $T^k_t$ is a superposition of a set of unit words $\{|w^1_t\rangle, |w^2_t\rangle,...,|w^n_t\rangle\}$, the superposition state of $T^k_t$ is formulated as:
\begin{small}
	\begin{align}
		|T^{k}_{t}\rangle = \sum_{j=1}^{n}z^{j}_{t}|w^{j}_{t}\rangle,~~~z^{j}_{t} = r^j_{t}e^{i\theta^{j}_t}
            \label{eq:complex}
	\end{align}
\end{small}
where $n$ is number of words in the $k^{th}$ target utterance.
$z^j_t$ is the probability amplitude expressed in the complex polar form. 
In QP, the complex probability amplitude depicts the position of a particle. $i$ in the probability amplitude is the imaginary number satisfying $i^2=-1$ , $r^j_t$ and $\theta^{j}_{t}\in \left [ -\pi ,\pi  \right ]$ represent amplitude and phase of $z^j_t$. We associate the amplitude $\textbf{R}$ and phase $\mathbf{\Theta}$ with specific linguistic meanings. The amplitude is analogous to the semantic knowledge. As for the phase, it is linked with the preassigned sentiment orientation of the utterance. The detailed explanation is provided in Appendix A.

Now, we obtain the complex-valued representation of the $k^{th}$ target utterance, namely 
$|T^{k}_{t}\rangle = \left ( r^{1}_{t}e^{i\theta^{1}_{t}},r^{2}_{t}e^{i\theta^{2}_{t}},...,r^{n}_{t}e^{i\theta^{n}_{t}} \right )^T$. 

For video, the low level visual features, e.g., visual sub-regions, can be seen as the basis units, which construct the visual Hilbert space $\mathcal{H}_v$. Thus, the visual counterpart of the $k^{th}$ utterance could be represented as: $|T^{k}_{v}\rangle = \left ( r^{1}_{v}e^{i\theta^{1}_{v}},r^{2}_{v}e^{i\theta^{2}_{v}},...,r^{n}_{v}e^{i\theta^{n}_{v}} \right )^T$.

For speech, we adopt the similar manner to treat the low level 
acoustic features, e.g., volume, frequency, as the basic units. We assume that the acoustic Hilbert space $\mathcal{H}_{a}$ is spanned by a set of orthogonal basis audio features $\{|w^{j}_{a}\rangle\}^{n}_{j=1}$, where the target speech can be written as: $|T^{k}_{a}\rangle = \left ( r^{1}_{a}e^{i\theta^{1}_{a}},r^{2}_{a}e^{i\theta^{2}_{a}},...,r^{n}_{a}e^{i\theta^{n}_{a}} \right )^T$.

\textbf{Contextual utterance representation.} The textual, visual and acoustic representations, i.e. $|C^k_{t}\rangle$ , $|C^k_{v}\rangle$, $|C^k_{a}\rangle$, of the $i^{th}$ context for the $k^{th}$ target utterance could be also calculated in the same way, i.e., Eq.~\ref{eq:complex}.

\subsection{Learning Intra-modality Contextuality with the Quantum Composition Layer}
Quantum composition describes the interaction between a quantum system and their surrounding environments. We treat the target utterance (or video, speech) as a quantum system, its context as the surrounding environment. We thus design a quantum composition layer to learn the intra-modality contextuality.

For text, given that the target utterance $|T^k_t\rangle$ and its contexts
$\{|C^1_t\rangle, |C^2_t\rangle ,..., |C^n_t\rangle\}$, a conversation sequence could be obtained as $\left \{Q^k = |C^1_t\rangle, |C^2_t\rangle ,..., |C^n_t\rangle, |T^k_t\rangle \right \}$. We feed these $n+1$ vectors in sequence $Q^k$ into a gate recurrent unit (GRU) network to produce their short contextual representations, we use hidden state generated at every step as current contextual feature, then we get $\{H^1_t, H^2_t,...,H^n_t, H^{n+1}_t\}$. In order to capture both long and short range contextual interactions, we represent the target utterance as a textual density matrix $\rho_{text}$, by encapsulating the outer product of each contextual representation. The density matrix has encoded all the information and interactions of utterance, which is computed as: 
\begin{small}
	\begin{align}
		\rho_{text} = \sum_{\lambda =1}^{n+1}p_\lambda |H^{\lambda}_t\rangle \langle H^{\lambda}_t|
		\label{eq:density}
	\end{align}
\end{small}
where $p_\lambda$ denotes the weight of interaction of each contextual representation. The density matrix $\rho_{text}$ encodes all information from the target utterance and its contexts.

For video and speech, two kinds of contextual representations are obtained via two separate GRUs, i.e., $\{H^1_v, H^2_v,...,H^n_v, H^{n+1}_v\}$ and $\{H^1_a, H^2_a,...,H^n_a, H^{n+1}_a\}$. Thus, two density matrices $\rho_{img}$ and $\rho_{auc}$ are also calculated using Eq.~\ref{eq:density}.

We obtain three density matrices $\rho_{text}$, $\rho_{img}$ and $\rho_{auc}$ for the target multi-modal sample. We feed them into the quantum interference-like inter-modal fusion layer for multi-modal fusion.

\subsection{Quantum Interference-like Fusion Layer}
We elaborate an analogy to quantum interference phenomenon in multi-modal fusion. The subjective attitude of the author is uncertain, which can be analogized as the particle's state. Two modalities, e.g., textual/visual, textual/acoustic and visual/acoustic are analogized as two paths. Bi-modal fused features could be seen as the probability distribution of the particle going through two paths. The information from each modality contributes to the final bi-modal features contemporaneously. Then we can model the modality interference via quantum interference.

\begin{figure}[t]
	\centering
        \scalebox{0.85}{
	\includegraphics[height=3in, width=2.4in]{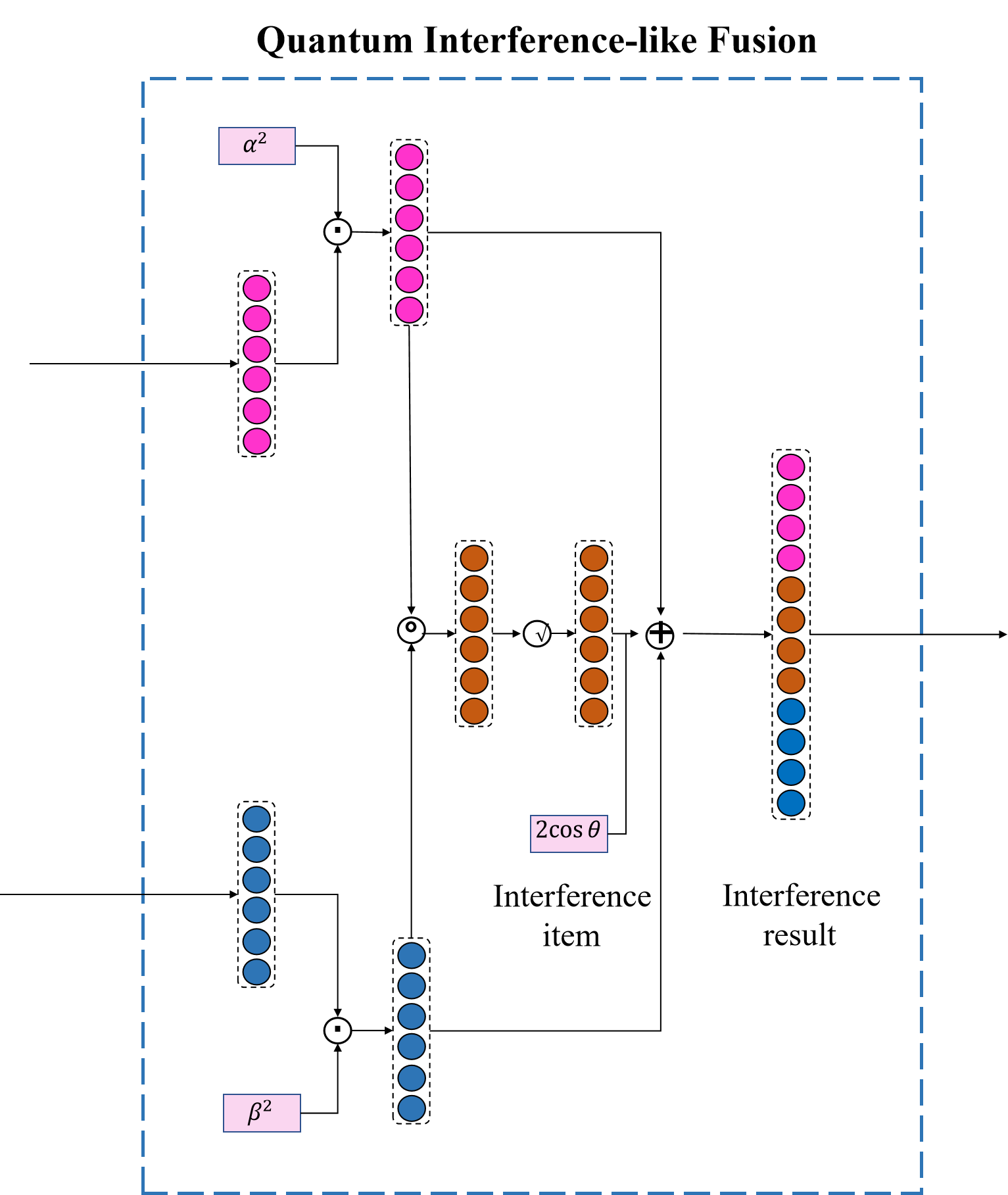}}
	\caption{Quantum interference-like fusion component. $\odot$ denotes point-wise multiplication. $\oplus$ refers to a element-wise addition. $\circledcirc$ is the matrix multiplication. \textcircled{$\sqrt{}$} refers to the square root operation.}
	\label{fig:interference-like fusion component}
\end{figure}

Based on Eq.~\ref{eq:eq5}, Eq.~\ref{eq:eq6} and Eq.~\ref{eq:eq7}, we argue that the subjective attitude of the speaker is in a quantum superposition-like of bi-modal representation, which can be expressed as:
\begin{equation}  \small
	\begin{aligned}
		z_p (x)= \alpha z_a (x) + \beta z_b (x) 
	\end{aligned}
\end{equation}
where $z_a (x)$\&$z_b (x)$ denotes the complex probability amplitudes of text-video, text-audio and video-audio pairs, e.g., $t\&v$, $a\&t$, $v\&a$ respectively. $z_p (x)$ denotes the complex probability amplitude of bi-modality. $f_a(x)=|\alpha|^2|z_a(x)|^2$ and $f_b(x)=|\beta|^2|z_b(x)|^2$ represent the corresponding probability distributions.
The probability distribution of bi-modal representation of the target document is written as:
\begin{equation}  \small
	\begin{aligned}
		f_p(x_k)
		=f_a(x_k)+f_b(x_k)+2\sqrt{f_a(x_k) f_b(x_k)}\cos{\phi_i}
	\end{aligned}
\end{equation}
where $x_k$ represents the $k^{th}$ feature component of the bi-modal representation $|f_p\rangle$, $I=2\sqrt{f_a(x) f_b(x)}\cos{\phi_i}$ is the interference item, capturing the non-linear interaction between different modalities, as shown in Figure~\ref{fig:interference-like fusion component}. 

We could get three bi-modal representations from $t\&v$, $a\&t$, $v\&a$, i.e., $|f_{tv}\rangle$, $|f_{ta}\rangle$ and $|f_{va}\rangle$. The final tri-modal representation $|f_{tva}\rangle$ is obtained by merging them together:
\begin{equation}  \small
	\begin{aligned}
	|f_{tva}\rangle = \left [|f_{tv}\rangle; |f_{va}\rangle; |f_{ta}\rangle \right ]
	\end{aligned}
\end{equation}

\subsection{Quantum Incompatible Measurement}
In QP, the information and property of a system (e.g., the author's sarcastic attitude) could be depicted by the probability distribution from the measurement outcomes. 
We perform a sequence of quantum incompatible measurements on tri-modal representation $|f_{tva}\rangle$, for obtaining the final sarcastic, sentimental and emotional features $\vec{m}^{sar}=\left ( m_1^{sar},m_2^{sar},...,m_G^{sar} \right )$, $\vec{m}^{sen}=\left ( m_1^{sen},m_2^{sen},...,m_G^{sen} \right )$ and $\vec{m}^{emo}=\left ( m_1^{emo},m_2^{emo},...,m_G^{emo} \right )$.

Three sets of measurement operators
$M^{sar}=\{M^{sar}_{\psi}\}^G_{\psi=1}$, $M^{sen}=\{M^{sen}_{\psi}\}^G_{\psi=1}$ and $M^{emo}=\{M^{emo}_{\psi}\}^G_{\psi=1}$
are constructed by performing the outer product of corresponding measurement vector $|D_{\psi}\rangle$, $|A_{\psi}\rangle$ and $|R_{\psi}\rangle$, that is 
$M^{sar}_{\psi}=|D_{\psi}\rangle \langle D_{\psi}|$,
$M^{sen}_{\psi}=|A_{\psi}\rangle \langle A_{\psi}|$,
$M^{emo}_{\psi}=|R_{\psi}\rangle \langle R_{\psi}|$.
The probability distribution after the measurement is written as:
\begin{equation}  \small
	\begin{aligned}
\vec{m}^{s}=tr\left ( \left ( M^{s} \right )^{\dagger }M^{s}  |f_{tva}\rangle\langle f_{tva}| \right )
	\end{aligned}
\end{equation}
where $s\in \left \{ sar, sen, emo \right \}$.
 
\subsection{Dense Layer}
The sarcastic, sentimental and emotional features $\vec{m}^{sar}$, $\vec{m}^{sen}$, $\vec{m}^{emo}$ are passed to the fully connected layer for each task respectively. The outputs are forwarded through the softmax functions to yield the sarcasm, sentiment, and emotion labels. We use cross entropy with L2 regularization as the loss functions $\zeta_{sar}$, $\zeta_{sen}$ and $\zeta_{emo}$ for training each task.
\begin{equation} 
	\begin{aligned}
		\zeta_{\gamma}
		= - 1/N \sum_{i}\sum_{n=1}^{E}y_{i}log(p_i)
	\end{aligned}
\end{equation}
where $\zeta_{\gamma}\in \left \{\zeta_{sar},\zeta_{sen},\zeta_{emo}\right \}$, $N$ is the number of samples in the dataset, $E$ denotes the number of classes, in this work $E_{sar}=2$, $E_{sen}=3$, $E_{emo}=9$, $y_i$ is the ground truth and $p_i$ is the prediction.

We jointly optimize three loss functions with different weights, which is written as:
\begin{equation} 
	\begin{aligned}
		\zeta = w_{sar}\zeta_{sar} + w_{sen}\zeta_{sen} + w_{emo}\zeta_{emo}
	\end{aligned}
\end{equation}

where $w_{sar}$, $w_{sent}$ and $w_{emo}$ satisfying $w_{sar}+w_{sen}+w_{emo}=1$. The dropout strategy is applied in the training stage to avoid overfitting.

\section{Experiments and Analysis}

\begin{table}[t]
\centering
\scalebox{0.9}{
\begin{tabular}{ccccc}
\hline
\multicolumn{1}{|c|}{\textbf{Dataset}}                  & \multicolumn{1}{c|}{\textbf{Task}}                  & \multicolumn{1}{c|}{\textbf{Classes}} & \multicolumn{1}{c|}{\textbf{No. of Utter.}} & \multicolumn{1}{c|}{\textbf{RC(\%)}} \\ 

\hline
\hline

\multicolumn{1}{|c|}{\multirow{16}{*}{\textbf{\textit{MUStARD$_{ext}$}}}} & 
\multicolumn{1}{c|}{\multirow{2}{*}{Sarcasm}} & \multicolumn{1}{c|}{Sar.} &
\multicolumn{1}{c|}{345} &
\multicolumn{1}{c|}{50.00} \\ 
\cline{3-5} 

\multicolumn{1}{|c|}{} & 
\multicolumn{1}{c|}{}  &
\multicolumn{1}{c|}{Non.} &
\multicolumn{1}{c|}{345} &
\multicolumn{1}{c|}{50.00} \\ 
\cline{2-5}

\multicolumn{1}{|c|}{} & \multicolumn{1}{c|}{\multirow{3}{*}{Sentiment}} & \multicolumn{1}{c|}{Pos.} &
\multicolumn{1}{c|}{210} &
\multicolumn{1}{c|}{30.43} \\ 
\cline{3-5} 

\multicolumn{1}{|c|}{} & 
\multicolumn{1}{c|}{}  &
\multicolumn{1}{c|}{Neg.} & 
\multicolumn{1}{c|}{391} & 
\multicolumn{1}{c|}{56.67} \\ 
\cline{3-5} 

\multicolumn{1}{|c|}{} & 
\multicolumn{1}{c|}{}  & 
\multicolumn{1}{c|}{Neu.} &
\multicolumn{1}{c|}{89} &
\multicolumn{1}{c|}{12.90} \\ 
\cline{2-5}

\multicolumn{1}{|c|}{} &
\multicolumn{1}{c|}{\multirow{9}{*}{Emotion}} & \multicolumn{1}{c|}{An.} &
\multicolumn{1}{c|}{97} &
\multicolumn{1}{c|}{14.05} \\
\cline{3-5} 

\multicolumn{1}{|c|}{} & 
\multicolumn{1}{c|}{}  &
\multicolumn{1}{c|}{Ex.} &
\multicolumn{1}{c|}{18} &
\multicolumn{1}{c|}{2.61} \\ 
\cline{3-5} 

\multicolumn{1}{|c|}{} & 
\multicolumn{1}{c|}{}  &
\multicolumn{1}{c|}{Fr.} &
\multicolumn{1}{c|}{14} &
\multicolumn{1}{c|}{2.03} \\ 
\cline{3-5} 

\multicolumn{1}{|c|}{} &
\multicolumn{1}{c|}{} &
\multicolumn{1}{c|}{Sd.} &
\multicolumn{1}{c|}{121} &
\multicolumn{1}{c|}{17.54} \\ 
\cline{3-5} 

\multicolumn{1}{|c|}{} & 
\multicolumn{1}{c|}{} &
\multicolumn{1}{c|}{Sp.} &
\multicolumn{1}{c|}{29} &
\multicolumn{1}{c|}{4.20} \\ 
\cline{3-5} 

\multicolumn{1}{|c|}{} &
\multicolumn{1}{c|}{} & 
\multicolumn{1}{c|}{Fs.} &
\multicolumn{1}{c|}{57} &
\multicolumn{1}{c|}{8.26} \\ 
\cline{3-5} 

\multicolumn{1}{|c|}{} &
\multicolumn{1}{c|}{} & 
\multicolumn{1}{c|}{Hp.} & 
\multicolumn{1}{c|}{143} & 
\multicolumn{1}{c|}{20.72} \\ 
\cline{3-5} 

\multicolumn{1}{|c|}{} & 
\multicolumn{1}{c|}{} & 
\multicolumn{1}{c|}{Neu.} & 
\multicolumn{1}{c|}{198} & 
\multicolumn{1}{c|}{28.70} \\ 
\cline{3-5} 

\multicolumn{1}{|c|}{} & 
\multicolumn{1}{c|}{}  & 
\multicolumn{1}{c|}{Dg.} &
\multicolumn{1}{c|}{39} & 
\multicolumn{1}{c|}{5.65} \\ 

\hline         
\hline

\multicolumn{1}{|c|}{\multirow{10}{*}{\textbf{\textit{MELD}}}} & 
\multicolumn{1}{c|}{\multirow{3}{*}{Sentiment}} & \multicolumn{1}{c|}{Pos.} &
\multicolumn{1}{c|}{3088} &
\multicolumn{1}{c|}{22.53} \\ 
\cline{3-5} 

\multicolumn{1}{|c|}{} & 
\multicolumn{1}{c|}{}  &
\multicolumn{1}{c|}{Neg.} & 
\multicolumn{1}{c|}{4194} & 
\multicolumn{1}{c|}{30.52} \\ 
\cline{3-5} 

\multicolumn{1}{|c|}{} & 
\multicolumn{1}{c|}{}  & 
\multicolumn{1}{c|}{Neu.} &
\multicolumn{1}{c|}{6436} &
\multicolumn{1}{c|}{46.95} \\ 
\cline{2-5}

\multicolumn{1}{|c|}{} &
\multicolumn{1}{c|}{\multirow{9}{*}{Emotion}} & 
\multicolumn{1}{c|}{An.} &
\multicolumn{1}{c|}{1607} &
\multicolumn{1}{c|}{11.72} \\
\cline{3-5} 

\multicolumn{1}{|c|}{} & 
\multicolumn{1}{c|}{}  &
\multicolumn{1}{c|}{Dg.} &
\multicolumn{1}{c|}{361} &
\multicolumn{1}{c|}{2.63} \\ 
\cline{3-5} 

\multicolumn{1}{|c|}{} & 
\multicolumn{1}{c|}{}  &
\multicolumn{1}{c|}{Fr.} &
\multicolumn{1}{c|}{358} &
\multicolumn{1}{c|}{2.61} \\ 
\cline{3-5} 

\multicolumn{1}{|c|}{} &
\multicolumn{1}{c|}{} &
\multicolumn{1}{c|}{Jy.} &
\multicolumn{1}{c|}{2308} &
\multicolumn{1}{c|}{16.84} \\ 
\cline{3-5} 

\multicolumn{1}{|c|}{} & 
\multicolumn{1}{c|}{} &
\multicolumn{1}{c|}{Neu.} &
\multicolumn{1}{c|}{6436} &
\multicolumn{1}{c|}{46.95} \\ 
\cline{3-5} 

\multicolumn{1}{|c|}{} &
\multicolumn{1}{c|}{} & 
\multicolumn{1}{c|}{Sd.} &
\multicolumn{1}{c|}{1002} &
\multicolumn{1}{c|}{7.31} \\ 
\cline{3-5} 

\multicolumn{1}{|c|}{} &
\multicolumn{1}{c|}{} & 
\multicolumn{1}{c|}{Sp.} & 
\multicolumn{1}{c|}{1636} & 
\multicolumn{1}{c|}{11.94} \\ 
\cline{3-5}
\hline   
\end{tabular}
}
\caption{Dataset statistics.In MUStARD$_{ext}$ emotions including, An: Anger, Ex: Excited, Fr: Funny, Fr: Fear, Sd: Sad, Sp: Surprised, Fs: Frustrated, Hp: Happy, Neu: Neutral, Dg: Disgust. In MELD emotions including, An: Anger, Dg: Disgust, Fr: Fear, Jy: Joy, Neu: Neutral, Sd: Sad, Sp: Surprised. In MUStARD$_{ext}$ we count the labels of the last utterance, in MELD we count labels for every utterance.}\label{tab:data}
\end{table}

\begin{center}
\begin{table}[t] \centering
\scalebox{1}{
\begin{tabular}{|c|c|c|}
\hline
\textbf{Dataset}                  & 
\textbf{Partition}                  &
\textbf{Count of Dialogues}\\ 
\hline
\hline

\multirow{3}{*}{\textbf{\textit{MUStARD$_{ext}$}}} & Train & 550 (2297 utterances) \\ 
\cline{2-3} 

& Dev. & 70 (319 utterances)  \\ \cline{2-3} 

& Test & 70 (336 utterances) \\ \cline{2-3}

\hline
\hline

\multirow{3}{*}{\textbf{\textit{MELD}}} & Train & 1039 (9989 utterances) \\ 
\cline{2-3} 

& Dev. & 114 (1109 utterances)  \\  \cline{2-3} 

& Test & 280 (2610 utterances) \\  \cline{2-3} 

\hline

\end{tabular}
}
\caption{Partition of the MUStARD$_{ext}$ dataset and the MELD dataset. }\label{tab:data_partition}
\end{table}
\end{center}

\subsection{Experiment Settings}
\textbf{Dataset.} To carry out an empirical evaluation, we need to choose benchmark datasets that have textual, visual and acoustic modalities with all sentiment, emotion and sarcasm labels. To this end, only the extended version of MUStARD ($MUStARD_{ext}$ for short) meet these criteria. The original $MUStARD$ dataset is made up of 3.68 hours conversational video, which consists of 690 samples total of 3,000 utterances. Each sample is a conversation consisting of several utterances. The samples are collected from 4 TV Series i.e., Friends, The Big Bang Theory, The Golden Girls, and Sarcasmaholics Anonymous, and are manually annotated. Chauhan et al.~\cite{chauhan2020sentiment} extended this dataset to sentiment and emotion scenario and re-annotated sentiment and emotion labels. 

In addition, in order to evaluate the robustness of the proposed model, we also applied it to bi-modal bi-task scenarios. We conduct experiments on another large scale dataset that only contains the sentiment and emotion labels, i.e., MELD~\cite{poria2018meld}. MELD contains 13,708 utterances from 1433 dialogues of the Friends TV series. The utterances in each dialogue are annotated with one of three sentiments (positive, negative or neutral) and one of seven emotions (anger, disgust, fear, joy, neutral, sadness or surprise). Table~\ref{tab:data} shows detailed statistics.

In Table~\ref{tab:data_partition} we show the partition of two dataset, we split datasets at the granularity of dialogues, count of dialogues for Train set, Dev. set and Test set are listed in the table. 


\textbf{Evaluation metrics.} We adopt $precision$ (P), $recall$ (R) and $micro-F1$ (Mi-F1) as evaluation metrics in our experiments. 

\textbf{Pre-processing.} For textual information, we first clean all the texts by checking for illegible characters and correcting spelling mistakes automatically. 

\textbf{Hyper-parameters.} As TensorFlow used to build the model does not support the complex representation, we take the real part and the imaginary part of a complex number as two separative inputs. The real parts of textual, visual and acoustic counterparts are initialized with BERT, EfficientNet and VGGish respectively. The phases in the imaginary parts are initialized with the pre-assigned sentiments using BERT. 
Based on the sentiment polarity result from BERT, we initialize every position in the phase part with a random number in (-pi,0) when the sentiment polarity is negative. While, when the sentiment polarity is positive, every position in the phase part is initialized with a random number in (0,pi).
The quantum measurement operators are randomly initialized with the unit vector and are set to be trainable. We evaluate the model by trying different combinations of hyper-parameters, and the finally selected hyper-parameters lead to the best performance. The optimal hyper-parameters are listed in Table ~\ref{tab:hyper} \footnote{Hyper-parameters pools are shown in the Appendix D}.

\begin{table}[t]
	\small
	\centering
	\scalebox{0.8}{
		\begin{tabular}{|l|cc|}
			\hline
			\textbf{Hyper-parameters} &  \multicolumn{2}{|c|}{\textbf{MUStARD$_{ext}$}} \\ \hline
			\hline
			Text Embedding size & \multicolumn{2}{c|}{768}  \\ 
			Image Embedding size & \multicolumn{2}{c|}{2048}  \\ 
			Audio Embedding size & \multicolumn{2}{c|}{128}  \\ 
			Phase Embedding size & \multicolumn{2}{c|}{128}  \\ 
			Phase initialation range & \multicolumn{2}{c|}{$-\pi - \pi$}  \\
			Activations &  \multicolumn{2}{c|}{Relu}   \\ 
			Batch size & \multicolumn{2}{c|}{64}   \\
			Learning rate &  \multicolumn{2}{c|}{0.0075}  \\
			No. of measurement &  \multicolumn{2}{c|}{1000}  \\
			Dropout rate &  \multicolumn{2}{c|}{0.2}  \\ 
			 Epochs &  \multicolumn{2}{c|}{100 (with early-stop)}  \\ 
			Interference item $\cos \phi_i $& \multicolumn{2}{c|}{-0.3}  \\
			Interference coefficient ($\alpha ^{2},\beta ^{2}$) & \multicolumn{2}{c|}{(0.5,0.5)} \\
			No. GRU cells & \multicolumn{2}{c|}{(128)} \\
			\hline
		\end{tabular}
	}
	\caption{Model configurations.}\label{tab:hyper}
\end{table}

To obtain the optimal experiment results, we use the early-stop strategy, which will stop training when the performance no longer increases. The self-adjusted learning rate changes as the training process goes. We conduct our experiments via $Keras \ v2.2.5$ open library. The detailed experimental environment is shown in Table~\ref{tab:exp_env}.

\begin{table}[t]
	\small
	\centering
	\scalebox{0.84}{
		\begin{tabular}{|l|cc|}
			\hline
			\textbf{Experiment Environment} &  \multicolumn{2}{|c|}{\textbf{}} \\ \hline
			\hline
			Operating system & \multicolumn{2}{c|}{Ubuntu 16.04.6}  \\ 
			GPU & \multicolumn{2}{c|}{GeForce RTX 2080ti}  \\
			GPU driver &  \multicolumn{2}{c|}{CUDA 11.2}   \\ 
			Memory size & \multicolumn{2}{c|}{10GB}  \\  
			Deep-learning framework & \multicolumn{2}{c|}{keras v2.2.5}  \\  
			$ \ $ & \multicolumn{2}{c|}{tensorflow-gpu v1.14.0}  \\ \hline
		\end{tabular}
	}
	\caption{Hardware devices and software environment.}\label{tab:exp_env}
\end{table}

\subsection{Baselines}
In this work, we treat sarcasm detection task as the main task, and select sentiment analysis and emotion recognition as the auxiliary tasks. We compare our QUIET model with a range of state-of-the-art baselines. They are listed as follows:

\textbf{SVM+BERT \cite{devlin2019bert}:} It uses BERT to represent textual utterances with the standard hyper-parameter settings. Besides, the kernel function for SVM is set to “RBF”. We also concatenate the contextual features.

\textbf{RCNN-RoBERTa \cite{potamias2020transformer}:} It utilizes pre-trained RoBERTa vectors as textual representation, and combines with a RCNN to capture contextual information.

\textbf{EfficientNet \cite{tan2019efficientnet}:} It uses a compound scaling method to create different models, which has achieved state-of-the-art performance on the ImageNet challenge.

\textbf{UPB-MTL \cite{vlad2020upb}:} It is a multi-modal multi-task learning architecture that combines ALBERT for text encoding with VGG-16 for image representation.

\textbf{QMSA \cite{zhang2018quantum}:} It first extracts visual and textual features using density matrices, and feeds them into the SVM classifier.

\textbf{A-MTL framework \cite{chauhan2020sentiment}:} It proposes an attention based multi-task model to simultaneously analyse sentiment, emotion and detect sarcasm.

\textbf{LF-DNN \cite{ding2022multimodal}:} It proposes a a multi-modal fusion model with residual connections based on late fusion.

\textbf{ConAttSD \cite{zhang2021multi}:} It constructs a contrastive-attention-based sarcasm detection (ConAttSD) model, and uses an inter-modality contrastive attention mechanism to extract the contrastive features for an utterance.

\textbf{Hybrid \cite{bharti2022multimodal}:} Designs a LSTM network based acoustic encoder and a CNN network based textual encoder to extract corresponding features. Two features are combined and then the hybrid feature is send into a SVM model to get classification result.

\textbf{KAMT \cite{zhang2021affective}:} It designs an external knowledge enhanced multi-task representation learning network, termed KAMT, for emotion recognition.

\textbf{Parameters analysis and complexity.} In Appendix Table.1, we present and compare the number of parameters from thirteen models, including the proposed QUIET and its variant QUIET-Double-Real, eight baseline models and other three pre-trained language models.


From Table.1 in Appendix, we notice that all baseline models have fewer parameters than pre-trained language models (i.e., ALBERT-base, BERT-base, BERT-large). And among the proposed model and all baseline models, UPB-MTL and EfficientNet have more parameters than QUIET, this is because they have the most complex structures among all baselines. Compared with ConAttSD, A-MTL and RCNN-RoBERTa, QUIET has about twice numbers of parameters. As for Hybrid and QMSA, QUIET has about four times the number of parameters. The main reason QUIET has more parameters is that it is based on the complex representation which contains real-part and complex-part thus makes the quantity of parameters doubled. However, our proposed model outperforms other baselines on three tasks with a considerable training time. We argue that our QUIET model make a good balance between time complexity and the classification performance.

\begin{table}[t]
	\small
	\begin{center}
		\resizebox{0.45\textwidth}{!}{\begin{tabular}{|c|l|c|c|c|}
				\hline
				\multirow{2}*{\textbf{Dataset}} &  \multirow{2}*{\textbf{Method}} & \multicolumn{3}{c|}{\textbf{Sarcasm Detection}} \\ 
				\cline{3-5}
				&  & P & R & $\mathrm{M_{i}}$-F1  \\ \hline
				\hline
				\multirow{11}*{\textbf{MUStARD$_{ext}$}} 
				& SVM+BERT & 65.14 & 64.61 & 64.68  \\
				& SVM+BERT (+context) & 65.53 & 65.11 &65.06 \\
				& RCNN-RoBERTa  & 68.70 & 64.33 &65.16 \\
				& EfficientNet   & 63.58 & 64.19 &63.77   \\
				& UPB-MTL   & 65.12 & 65.41 & 65.41  \\
				& QMSA   & 70.23 & 70.04 & 70.00  \\
				& Hybrid & 75.11 & 66.28 & 70.35 \\
				& LF-DNN & 73.82 & 73.77 & 73.75 \\
				& ConAttSD & 74.46 & 74.01 & 73.97 \\ 
				& A-MTL & 77.09 & 76.67 & 76.57 \\\cline{2-5}
				& Text-QUIET  & 72.44 & 71.45 & 72.13  \\
				& Image-QUIET  & 71.89 & 71.33 & 70.89  \\ 
				& Audio-QUIET  & 78.91 & 77.50 & 78.14   \\\cline{2-5}
				&   \textbf{QUIET}  &\textbf{83.74} & \textbf{82.93} & \textbf{83.70}  \\ 
				& $\bigtriangleup$SOTA & (+6.7\%)  & (+6.3\%) &  (+7.1\%)  \\\hline\hline
				
				\multirow{2}*{\textbf{Dataset}} &  \multirow{2}*{\textbf{Method}} & \multicolumn{3}{c|}{\textbf{Sentiment Analysis}} \\ 
				\cline{3-5}
				&  & P & R & $\mathrm{M_{i}}$-F1  \\ \hline
				\hline
				\multirow{11}*{\textbf{MUStARD$_{ext}$}} 
				& SVM+BERT & 57.17 & 57.10 & 57.23   \\
				& SVM+BERT (+context) & 58.44 & 58.21 & 58.41  \\
				& RCNN-RoBERTa  & 59.45 & 59.74 & 59.28  \\
				& EfficientNet   & 59.12 & 58.87 & 59.19        \\
				& UPB-MTL   & 59.81 & 59.72 & 59.36   \\
				& QMSA   & 59.67 & 60.05 & 59.78   \\
				& A-MTL & 60.56 & 60.69 & 61.17  \\\cline{2-5}
				& Text-QUIET  & 73.24 & 73.39 & 73.55  \\
				& Image-QUIET  & 66.52 & 66.81 & 66.70   \\ 
				& Audio-QUIET  & 63.72 & 63.63 & 63.89    \\\cline{2-5}
				&   \textbf{QUIET}  &\textbf{74.37} & \textbf{74.56} & \textbf{74.89}  \\ 
				& $\bigtriangleup$SOTA & (+13.8\%)  & (+13.9\%) &  (+13.7\%)  \\\hline\hline

				\multirow{2}*{\textbf{Dataset}} &  \multirow{2}*{\textbf{Method}} & \multicolumn{3}{c|}{\textbf{Emotion Recognition}} \\ 
				\cline{3-5}
				&  & P & R & $\mathrm{M_{i}}$-F1  \\ \hline
				\hline
				\multirow{11}*{\textbf{MUStARD$_{ext}$}} 
				& SVM+BERT & 25.64 & 25.71 & 25.67  \\
				& SVM+BERT (+context) & 26.39 & 26.39 & 26.39  \\
				& RCNN-RoBERTa  & 32.61 & 32.70 & 32.65  \\
				& EfficientNet  & 31.43 & 31.47 & 31.44   \\
				& UPB-MTL   & 33.55 & 33.64 & 33.60  \\
				& QMSA   & 24.14 & 24.53 & 25.10  \\
				& A-MTL  & 33.12 & 33.07 & 33.10 \\
				& KAMT   & 33.72 & 33.80 & 33.80  
				\\\cline{2-5}
				& Text-QUIET  & 34.23 & 34.17 & 34.20  \\
				& Image-QUIET  & 36.14 & 36.10 & 36.15 \\ 
				& Audio-QUIET  & 31.90 & 31.62 & 31.76  \\\cline{2-5}
				&   \textbf{QUIET}  &\textbf{37.64} & \textbf{37.71} & \textbf{37.69}  \\ 
				& $\bigtriangleup$SOTA & (+4.5\%)  & (+4.6\%) &  (+4.6\%)
				\\\hline
		\end{tabular}}
	\end{center}
	\caption{\label{tab-mustard} Comparison of different baseline models on three tasks on MUStARD$_{ext}$ dataset.}
\end{table}

\begin{table}[t]
	\small
	\begin{center}
		\resizebox{0.455\textwidth}{!}{
		\begin{tabular}{|c|l|c|c|c|}
				\hline
				\multirow{2}*{\textbf{Dataset}} &  \multirow{2}*{\textbf{Method}} & \multicolumn{3}{c|}{\textbf{Sentiment Analysis}} \\ 
				\cline{3-5}
				&  & P & R & $\mathrm{M_{i}}$-F1  \\ \hline
				\hline
				\multirow{7}*{\textbf{MELD}} 
				& SVM+BERT (+context) & 63.79 & 62.41 & 63.12  \\
				& EfficientNet   & 62.87 & 62.36 & 62.52        \\
				& UPB-MTL   & 60.94 & 61.47 & 61.04   \\
				& RCNN-RoBERTa & 62.48 & 60.32 & 61.42 \\
				& QMSA   & 65.21 & 65.87 & 65.33   \\\cline{2-5}
				&   \textbf{QUIET}  &\textbf{67.93} & \textbf{67.31} & \textbf{67.41}  \\ 
				& $\bigtriangleup$SOTA & (+2.72\%)  & (+1.44\%) &  (+2.08\%)  \\\hline\hline

				\multirow{2}*{\textbf{Dataset}} &  \multirow{2}*{\textbf{Method}} & \multicolumn{3}{c|}{\textbf{Emotion Recognition}} \\ 
				\cline{3-5}
				&  & P & R & $\mathrm{M_{i}}$-F1  \\ \hline
				\hline
				\multirow{7}*{\textbf{MELD}} 
				& SVM+BERT (+context) & 33.09 & 33.31 & 33.20  \\
				& EfficientNet  & 34.43 & 33.97 & 34.13   \\
				& UPB-MTL   & 33.21 & 33.54 & 33.38  \\
				& RCNN-RoBERTa & 34.53 & 33.91 & 33.69 \\
				& QMSA   & 36.01 & 37.37 & 36.56 \\\cline{2-5}
				&   \textbf{QUIET}  &\textbf{42.56} & \textbf{41.67} & \textbf{41.88}  \\ 
				& $\bigtriangleup$SOTA & (+2.13\%)  & (+1.7\%) &  (+1.74\%)
				\\\hline
		\end{tabular}}
	\end{center}
	\caption{\label{tab-meld} Comparison of different baseline models on two tasks on MELD dataset.}
\end{table}

\subsection{Comparative Analysis}
The experimental results are summarized in Table~\ref{tab-mustard}. We can notice that in the cases of sarcasm detection, the two popular pre-trained
language models, EfficentNet and SVM+BERT perform poorly among all baselines, and get the worst results. The reason is that we only fine-tune both models instead of improving their architectures. Through taking the conversational context into consideration, SVM+BERT (context) slightly outperforms the above-mentioned models for all three tasks, indicating that the conversation context would affect the sarcasm polarity of the target utterance. It is necessary to model the context. 
RCNN-RoBERTa performs better than SVM+BERT and EfficentNet. The major reasons are: (1) RNN could learn effective contextual information; and (2) RoBERTa is trained on a much larger dataset. However, it is only designed for text, which is not inapplicable to multi-modal learning. 
UPB-MTL outperforms SVM+BERT and other above-mentioned baselines for all three tasks. Because UPB-MTL is built on the top of two pre-trained models, e.g., BERT and ResNet, it can leverage the complementary information from the two models. 

QMSA performs well for the tasks of sentiment analysis and sarcasm detection, while performs poorly in the case of emotion recognition. The performance of it varies largely for different tasks. This may be due to the instability of quantum density matrix. Our QUIET model will improve the shortcomings of QMSA by designing an end-to-end quantum probability inspired framework. Hybrid obtains comparable results against QMSA. The reason is that Hybrid only adopts a simple fusion architecture. LF-DNN and ConAttSD outperform the above-mentioned baselines significantly. The reasons are: (1) the residual connection based late fusion prevents the degeneration; (2) the contrastive attention module could learn more complementary knowledge from multiple modalities. However, both of them are weaker than A-MTL, because A-MTL models the interaction across different tasks.

A-MTL performs well and achieves the best classification performance among all baselines for the tasks of sarcasm and sentiment analysis, and gets comparable results against UPB-MTL for the task of emotion recognition. 
Compared with UPB-MTL, the micro-f1 scores increase by 15.1\% and 3.7\%. Because it unifies pre-trained language models (PLM), multi-task learning and two attention mechanisms into a framework, which could better combine the information across the modalities to effectively classify sarcasm, sentiment, and emotion.

Text-QUIET, Image-QUIET and Audio-QUIET surpass SVM+BERT, RCNN-RoBERTa and UPB-MTL, but under-performs A-MTL. This result shows that the uni-modal setup of the proposed QUIET model can still achieve comparable performance against strong baselines. In this work, we will not treat them independently. Meanwhile, Text-QUIET has shown its best robustness against another two uni-modal setups. Finally, the proposed QUIET model achieves the best micro-F1 scores of 83.7\%, 74.89\%, 37.69\%
against micro-F1 scores of 76.57\%, 61.17\%, 33.1\% of the state-of-the-art baselines. This empirically proves the effectiveness and feasibility of QUIET, and its great potential in human affect analysis. We will conduct detailed analysis of QUIET from other aspects in the following sections.

In Table~\ref{tab-meld}, classification results on MELD dataset are listed. Among all tested baseline models, QMSA, which is a QP based baseline model, got the best result on both tasks. Comparing with QMSA, QUIET got improvement on both task. On sentiment analysis task, it is 2.72\%, 1.44\%, 2.08\% for precision, recall, and micro-f1 score. On emotion recognition task, improvements are 2.13\%, 1.7\%, and 1.74\% for three metrics. On the MELD dataset, our model shows the best result, which shows the effectiveness and great generalization of the proposed model.

In these experiments, emotion recognition is the most complex task in comparison to sentiment analysis and sarcasm detection. This is because sarcasm detection task is a binary classification task, and sentiment analysis is a ternary classification task. For emotion recognition, the class number expends from 2/3 to 9. Having such many categories is a direct reason for emotion recognition being such a complex task (The average probability of correctly predicting a label has increased from 1/2 or 1/3 to 1/9). Multi-class classification task is more difficult than binary class task. In addition, from the cognition perspective, it is difficult to distinguish the specific emotion from similar emotions, e.g., distinguishing the disgust from anger, since such different emotions do not have clear boundary.
From the quantum theory perspective, emotion state is under a superposition state composed by 9 different basis states, this leads to a smaller probability of collapsing into each emotion state compared to sarcasm and sentiment.

\textbf{Significance test.} 
We have employed the paired t-test to perform significance test on baseline models and ablation models. Results (p-value) are shown in Appendix Table.2 and Table.3.
We observe that the performance improvement in the proposed models over the state-of-the-art systems is significant with 95\% confidence (i.e., p-value< 0.05). 
In addition, we notice that the p-values of sarcasm detection task on tri-task-va v/s bi-task(sar+sent)-va, QUIET(sar) v/s uni-task(sar)-tri-modal are larger than 0.05. For the first case, we argue that decision on emotion recognition affects decisions on sentiment analysis and sarcasm, because both result in tri-task-va is lower than task(sar+sent)-va. And acoustic pre-trained model is not as efficient as textual and visual pre-trained model (the vggish model was transformed from the visual pre-trained model), this is also one of the reasons for this case. For the second case, QUIET(sar) works better than uni-task(sar)-tri-modal, however p-value is a little higher than 0.05, which is 0.06346. We argue this is because the distribution of predicted labels is more similar compared to the results of other models.

\subsection{Single-Task vs. Multi-Task Learning}
In order to analyze the role of multi-task learning, we depict the comparison results between the multi-task learning (MTL) and single-task learning (STL) frameworks in Table~\ref{tab:sml}. We compare the single task setup with bi-task and tri-task learning.

We can observe that the best F1 scores for single task setup are 74.44\%, 72.01\% and 31.13\% for sarcasm, sentiment, and emotion respectively. In a bi-task setup, the best scores for sarcasm, sentiment, and emotion are 77.56\%, 71.87\% and 32.25\%. We see that the performance of sarcasm detection and emotion recognition has improved via bi-task learning, while the performance of sentiment analysis is comparable. We also conclude that both sentimental and emotional knowledge help sarcasm detection, especially the former. The sentimental knowledge facilitates the identification of emotion. In a tri-task setup, the best F1 scores for sarcasm, sentiment, and emotion are 78.98\%, 73.55\% and 36.15\% respectively, which significantly outperform the results of all single task and bi-task setups. This shows the effectiveness of multi-task learning.

We have also performed another experiment to explore the impact of multi-modality on single task and multi-task setups. The experimental results are shown in Table~\ref{tab:bml} and Table~\ref{tab:tml}. We can observe that all F1 scores of multi-modal cases for all setups (including single task, bi-task and tri-task learning) significantly outperform that of uni-modality (e.g., Text, Video, Audio). For example, the F1 scores for the tri-modality case in the tri-task setting are 83.74\%, 74.89\% and 37.69\%, as compared to the best F1 scores of uni-modality, i.e., 78.98\% (with V), 73.55\% (using T) and 36.15\% (using V). The above results implicate that the importance of multi-task learning and multi-modal modeling, and our QUIET model has incorporated both of them into a unified framework.

\begin{table}[t]
	\begin{center}
			\small
		\scalebox{0.85}{
			\begin{tabular}{|l|l|c|c|c|}
				\hline
				\multirow{2}{*}{\textbf{Setup}} & \multirow{2}{*}{\textbf{Task}} & \multicolumn{1}{c|}{\textbf{T}} & \multicolumn{1}{c|}{\textbf{A}} & \multicolumn{1}{c|}{\textbf{V}} \\ \cline{3-5} 
				& & $\mathrm{M_{i}}$-F1 &  $\mathrm{M_{i}}$-F1 & $\mathrm{M_{i}}$-F1 \\ 
				\hline
				\hline
				
				\multirow{1}{*}{Single task} & Sarcasm & 70.33 &  68.97 & 74.44 \\ \cline{2-5} 
				\hline
				
				\multirow{1}{*}{Single task} & Sentiment & 71.97 &  66.73 & 72.01 \\ \cline{2-5} 
				\hline
				
				\multirow{1}{*}{Single task} & Emotion & 31.13 & 30.90 & 31 .00 \\ \cline{2-5} 
				\hline
				\hline
				
				\multirow{2}{*}{Sar+Emo (bi-task)} & Sarcasm & 71.25 &  69.28 & 77.33 \\
				& Emotion & 30.49 & 30.13 & 30.76 \\ \hline
				
				\multirow{2}{*}{Emo+Sent (bi-task)} & Emotion & 32.25 & 29.32 & 31.11 \\
				& Sentiment & 70.27 & 71.87 & 71.21 \\ \hline
				
				\multirow{2}{*}{Sent+Sar (bi-task)} & Sentiment &  70.77 & 65.47 & 71.20 \\
				&  Sarcasm & 71.18 & 69.56  & 77.56 \\ \hline
				
				\multirow{3}{*}{Sar+Sent+Emo (tri-task)}&  Sarcasm &  71.98 & 72.36 & 78.98 \\
				&  Sentiment &  73.55  & 63.89 & 66.70  \\ 
				&  Emotion & 34.20 &   31.76 & 36.15 \\ \hline
			\end{tabular}
		}
		\caption{\label{tab-sml} Comparison with single-task learning (STL) and multi-task (MTL) learning frameworks on three single modalities. T: Text, V: Visual, A: Audio}
		\label{tab:sml}
	\end{center}
\end{table}

\begin{table}[t]
	\begin{center}
			\small
		\scalebox{0.85}{
			\begin{tabular}{|l|l|c|c|c|}
				\hline
				\multirow{2}{*}{\textbf{Setup}} & \multirow{2}{*}{\textbf{Task}} & \multicolumn{1}{c|}{\textbf{T+A}} & \multicolumn{1}{c|}{\textbf{V+T}} & \multicolumn{1}{c|}{\textbf{V+A}} \\ \cline{3-5} 
				& & $\mathrm{M_{i}}$-F1 & $\mathrm{M_{i}}$-F1 & $\mathrm{M_{i}}$-F1  \\ 
				\hline
				\hline
				
				\multirow{1}{*}{Single task}
				& Sarcasm &  73.24  & 75.88 & 74.87 \\ \cline{2-5} 
				\hline
				
				\multirow{1}{*}{Single task} 
				& Sentiment &   72.12  &  73.45 & 72.23 \\ \cline{2-5} 
				\hline
				
				\multirow{1}{*}{Single task} 
				& Emotion & 33.49 &  33.64 & 33.07 \\ \cline{2-5} 
				\hline
				\hline
				
				\multirow{2}{*}{Sar+Emo (bi-task)}
				&  Sarcasm &  74.29 &  79.79  & 74.97 \\
				&  Emotion & 34.14  & 33.41 & 34.23 \\
				\hline
				
				\multirow{2}{*}{Emo+Sen (bi-task)} 
				&  Emotion &  33.48 & 34.33 & 30.76 \\
				&  Sentiment & 72.74 & 73.26 & 72.06 \\
				\hline
				
				\multirow{2}{*}{Sen+Sar (bi-task)} 
				&  Sentiment & 70.94 &   73.56 & 72.08 \\
				&  Sarcasm  & 71.98 & 78.04 & 82.98 \\
				\hline
				
				\multirow{3}{*}{Sar+Sen+Emo (tri-task)}& Sarcasm & 75.85 &  81.52 & 76.99 \\  
				&  Sentiment &  73.33 &   70.64 & 71.10 \\  
				&  Emotion & 34.65 & 34.83 & 33.37 \\
				\hline
			\end{tabular}
		}
		\caption{\label{tab-bml} Comparison with single-task learning (STL) and multi-task (MTL) learning frameworks on combination of two modalities. T: Text, V: Visual, A: Audio}
		\label{tab:bml}
	\end{center}
\end{table}

\begin{table}[t]
	\begin{center}
		\small
		\scalebox{0.85}{
			\begin{tabular}{|l|l|c|}
				\hline
				\multirow{2}{*}{\textbf{Task}} & \multirow{2}{*}{\textbf{Setups}} & \multicolumn{1}{c|}{\textbf{T+A+V}} \\ \cline{3-3}
				&  &  $\mathrm{M_{i}}$-F1 \\ 
				\hline
				\hline
				
				\multirow{1}{*}{Sarcasm}
				& Single task &  79.96  \\
				\hline
				
				\multirow{1}{*}{Sentiment}
				& Single task &   73.65 \\
				\hline
				
				\multirow{1}{*}{Emotion}
				& Single task & 34.13 \\ 
				\hline
				\hline
				
				\multirow{2}{*}{Sar+Emo}&  Sarcasm &  76.81 \\
				&  Emotion & 33.63   \\ \hline
				
				\multirow{2}{*}{Emo+Sent} &  Emotion &  32.05 \\   
				&  Sentiment & 72.44 \\ \hline
				
				\multirow{2}{*}{Sent+Sar}&  Sentiment &  72.10 \\
				&  Sarcasm  &  82.17 \\ \hline
				
				\multirow{3}{*}{Sar+Sent+Emo} &  Sarcasm &  83.74 \\ 
				&  Sentiment &  74.89   \\
				&  Emotion & 37.69  \\ \hline
			\end{tabular}
		}
		\caption{\label{tab-tml} Comparison with single-task learning (STL) and multi-task (MTL) learning frameworks on combination of all three modalities. T: Text, V: Visual, A: Audio}
		\label{tab:tml}
	\end{center}
\end{table}

\subsection{Ablation Study}
To study the effectiveness of different components of the QUIET model, we perform the ablation study. We choose to remove only one component at each time and evaluate its impact on the overall performance. Four sub-models are designed: (1) $QUIET-Real$ that does not consider the complex embedding, i,e., replacing utterance embeddings with their real counterparts only; (2) $QUIET-Real-Double-Para$ that doubles the real part of the complex representation to make the parameter quantity equals to the proposed model; (3) $QUIET-No-Context$ that does not model the contextuality; (4) $QUIET-Concat$ that replaces quantum interference fusion with a feature concatenation operation; (5) $QUIET-Trad$ that replaces quantum incompatible measurement with a traditional softmax layer.

The experimental results are shown in Table~\ref{tab:ablation}. We can see that all the sub-models under-perform the QUIET model for all of the three tasks. Among the sub-models, QUIET-No-Context performs the poorest for the task of sarcasm detection. The reason is that sarcasm understanding is more dependent on the context. For sentiment analysis and emotion recognition, QUIET-Real achieves the worst performance, which shows that the imaginary part of the complex-valued representation is quite crucial in term of leveraging the prior knowledge to improve the efficiency of representation learning. QUIET-No-Context gets a better classification performance over the other sub-models in the case of sentiment analysis. One possible reason is that detecting sentiment orientation mainly relies on the current utterance. For emotion recognition, QUIET-Trad obtains the best F1 score among all the sub-models, which shows that quantum incompatible measurement contributes less to emotion recognition than to sarcasm and sentiment. We design QUIET-Real-Double-Para, which doubles the real part of the complex representation to ensure that the amount of parameters equals to the QUIET model. Results show that the expansion on the parameters benefits to the performance, especially on the sentiment analysis and emotion recognition tasks. QUIET-Real-Double-Para overcomes QUIET-Real because of the increase in the dimension of real part vectors. However, it under-performs than QUIET-Trad and the standard QUIET model for three tasks. This shows that the single increase in the dimension of vectors is not a good way to improve the performance. The improvement of the model mainly comes from our proposed QP framework rather than the expansion of parameters. In summary, all baselines are weaker than the proposed QUIET model, which proves all quantum components have their contributions.

\begin{table}[t]
	\small
	\begin{center}
		\scalebox{0.83}{
			\begin{tabular}{|c|l|c|}
				\hline
				\multirow{2}*{\textbf{Task}} & \multirow{2}{0.2\textwidth}{\centering \textbf{Models}} & \multicolumn{1}{c|}{\textbf{Metrics}} \\ 
				\cline{3-3}
				& & $\mathrm{M_{i}}$-F1 \\ \cline{1-3}
				\hline
				\hline
				\multirow{6}*{Sarcasm} & QUIET-Real & 74.13 \\ 
				& QUIET-Real-Double-Para & 77.21 \\ 
				& QUIET-No-Context & 64.92  \\
				&  QUIET-Concat & 75.87 \\
				& QUIET-Trad & 78.51 \\
				\cline{2-3}
				& QUIET-Sarcasm & 83.74  \\
				\hline
				\hline
				\multirow{6}*{Sentiment} & QUIET-Real & 48.97 \\ 
				& QUIET-Real-Double-Para & 55.42 \\ 
				& QUIET-No-Context & 69.81 \\
				&  QUIET-Concat & 61.50 \\
				& QUIET-Trad & 65.76 \\
				\cline{2-3}
				& QUIET-Sentiment & 77.53 \\
				\hline
				\hline
				\multirow{6}*{Emotion} & QUIET-Real & 22.87 \\ 
				& QUIET-Real-Double-Para & 30.33 \\ 
				& QUIET-No-Context & 26.14 \\
				&  QUIET-Concat & 26.87 \\
				& QUIET-Trad & 35.16 \\
				\cline{2-3}
				& QUIET-Emotion & 37.69  \\
				\hline
			\end{tabular}
		}
	\end{center}
	\caption{\label{tab:ablation} Ablation experiment results.}
\end{table}

\subsection{Context Range Study}
In order to analyze the effect of context range, we calculate the distribution of different context ranges in the dataset, where detailed statistics are shown in Table~\ref{tab:context_number}. We notice a fact that 65\% utterances have less than three contextual utterances. Hence, we empirically set the upper limit of the context to three, and study the impact of different context ranges on the performance.

The results are reported in Tables~\ref{tab:range} with different context scopes. Zero context means that we only use the target utterance, ignoring the contextuality. One context utterance denotes that we use one history utterance before the target utterance to construct the density matrix. Two contexts mean that we use the previous two history utterances. And all context means we use all three previous contexts.

From Table~\ref{tab:range}, we observe that performance of all three tasks steadily increased, as context ranges increase. For example, the F1 scores are 64.92\%, 73.01\%, 80.23\% and 83.74\% respectively. This shows the important role of conversation contexts. QUIET with zero context expectantly performs the worst. QUIET with all contexts setup achieves the best F1 scores for all three tasks, which implies that taking all conversation contexts into consideration may be the best way to reach optimal performance.

\begin{table}[t]
	\small
	\begin{center}
                \scalebox{0.80}{
			\begin{tabular}{|c|c|}
					\hline
					\textbf{Context Range} & \textbf{No. of Utterance} \\ 
					\hline
					\hline
					1-3 & 445 \\
					4-7 & 197 \\
					8-12 & 48 \\
					\hline
					total & 690\\
					\hline
			\end{tabular}}
	\end{center}
	\caption{\label{tab:context_number} Counts of different context ranges.}
\end{table}

\begin{table}[t]
	\small
	\begin{center}
		\scalebox{0.78}{
			\begin{tabular}{|c|c|c|}
				\hline
				\multirow{2}*{\textbf{Task}} & \multirow{2}*{\textbf{Context Range}} & \multicolumn{1}{c|}{\textbf{Metrics}} \\ 
				\cline{3-3}
				& & $\mathrm{M_{i}}$-F1 \\ \hline
				\hline
				\multirow{4}*{Sarcasm} 
				& Zero & 64.92 \\
				&One & 73.01 \\
				& Two & 80.23 \\
				\cline{2-3}
				& All& 83.74  \\
				\hline
				\hline
				\multirow{4}*{Sentiment} 
				& Zero & 68.81 \\
				&One & 69.45 \\
				& Two & 71.64 \\
				\cline{2-3}
				& All& 77.53  \\
				\hline
				\hline
				\multirow{4}*{Emotion} 
				& Zero & 26.14 \\
				&One & 34.76\\
				& Two & 35.98 \\
				\cline{2-3}
				& All& 37.69 \\
				\hline
			\end{tabular}
		}
	\end{center}
	\caption{\label{tab:range} Effect of context range.}
\end{table}

\subsection{Error Analysis}
\begin{figure}[t]
No.1
\begin{minipage}[c]{0.63\textwidth}
\includegraphics[width=0.15\textwidth]{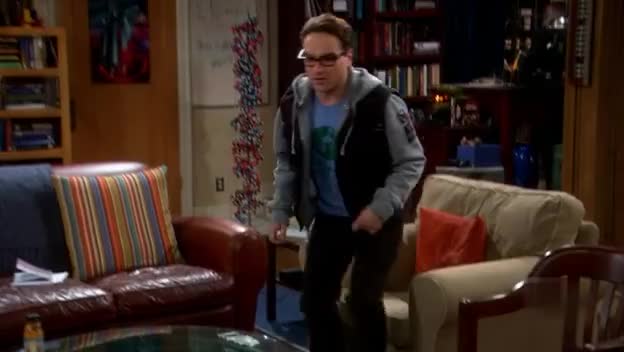}\vspace{1pt}
\includegraphics[width=0.15\textwidth]{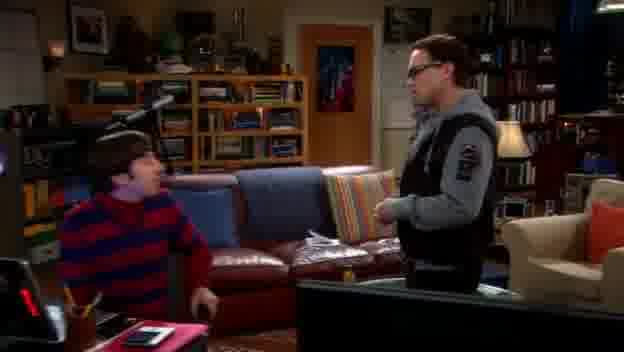}\vspace{1pt}
\includegraphics[width=0.15\textwidth]{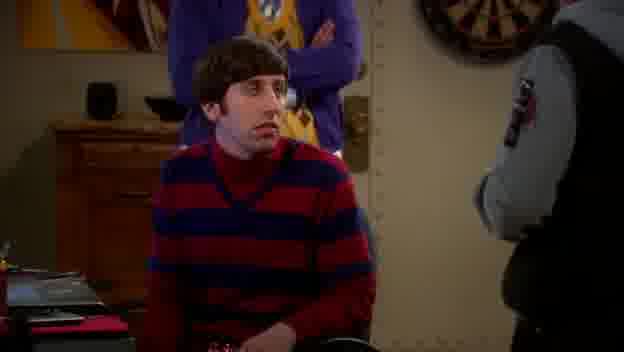}\vspace{1pt}
\includegraphics[width=0.15\textwidth]{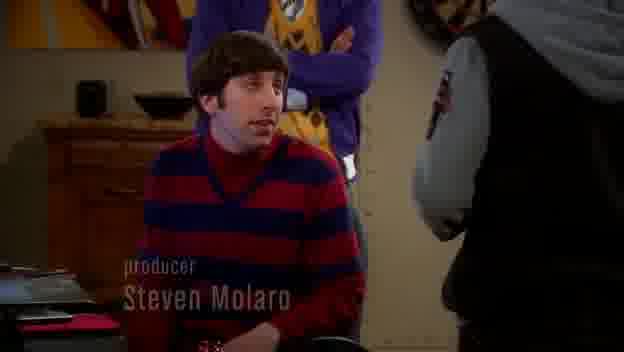}\vspace{1pt}
\end{minipage}

No.2
\begin{minipage}[c]{0.63\textwidth}
\includegraphics[width=0.15\textwidth]{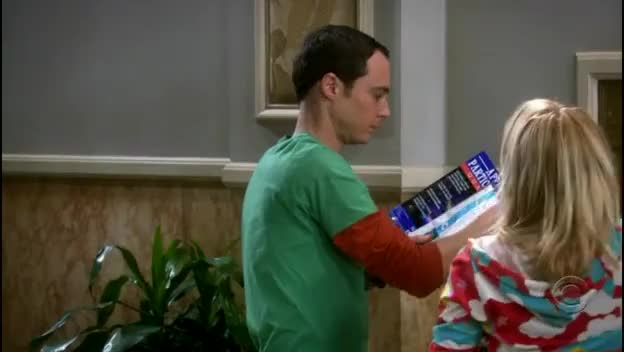}\vspace{1pt}
\includegraphics[width=0.15\textwidth]{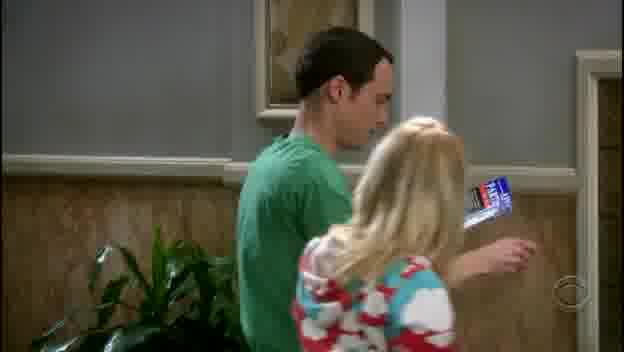}\vspace{1pt}
\includegraphics[width=0.15\textwidth]{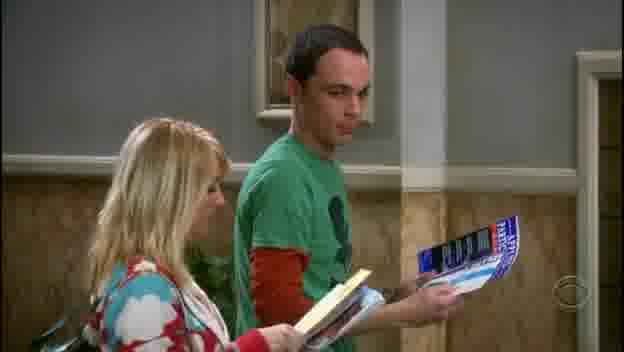}\vspace{1pt}
\includegraphics[width=0.15\textwidth]{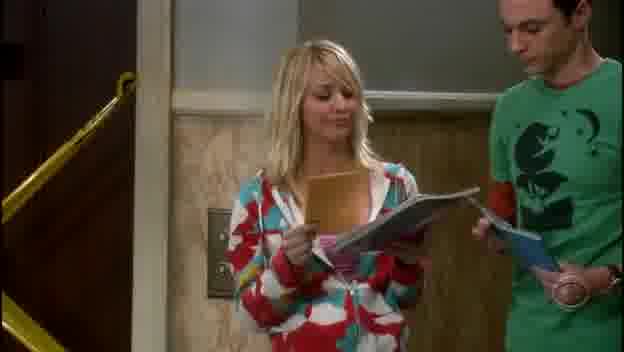}\vspace{1pt}
\end{minipage}

No.3
\begin{minipage}[c]{0.63\textwidth}
\includegraphics[width=0.15\textwidth]{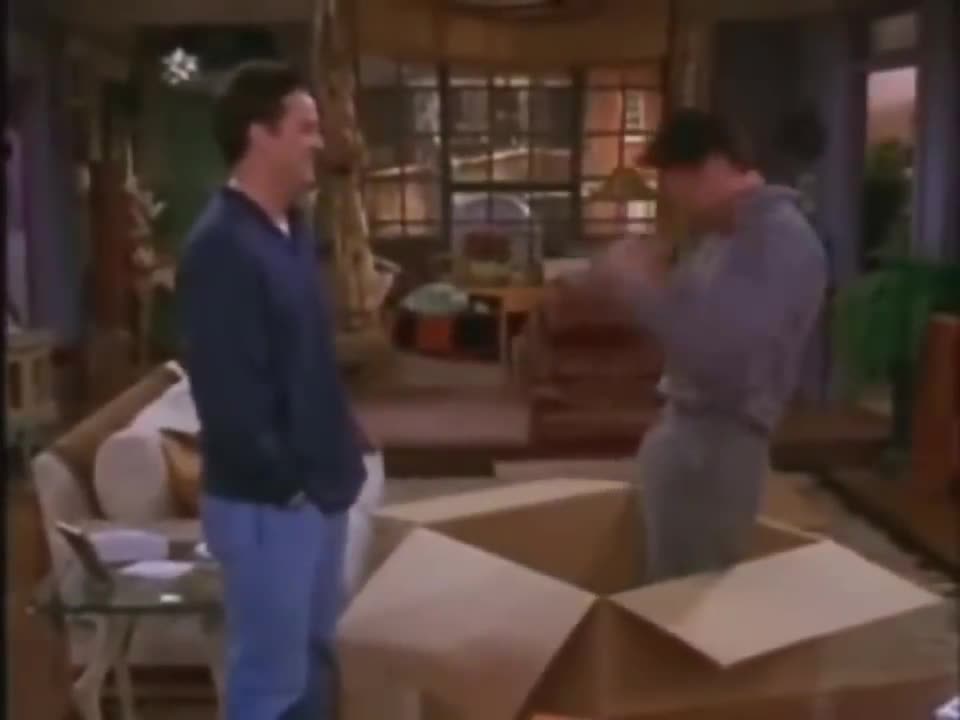}\vspace{1pt}
\includegraphics[width=0.15\textwidth]{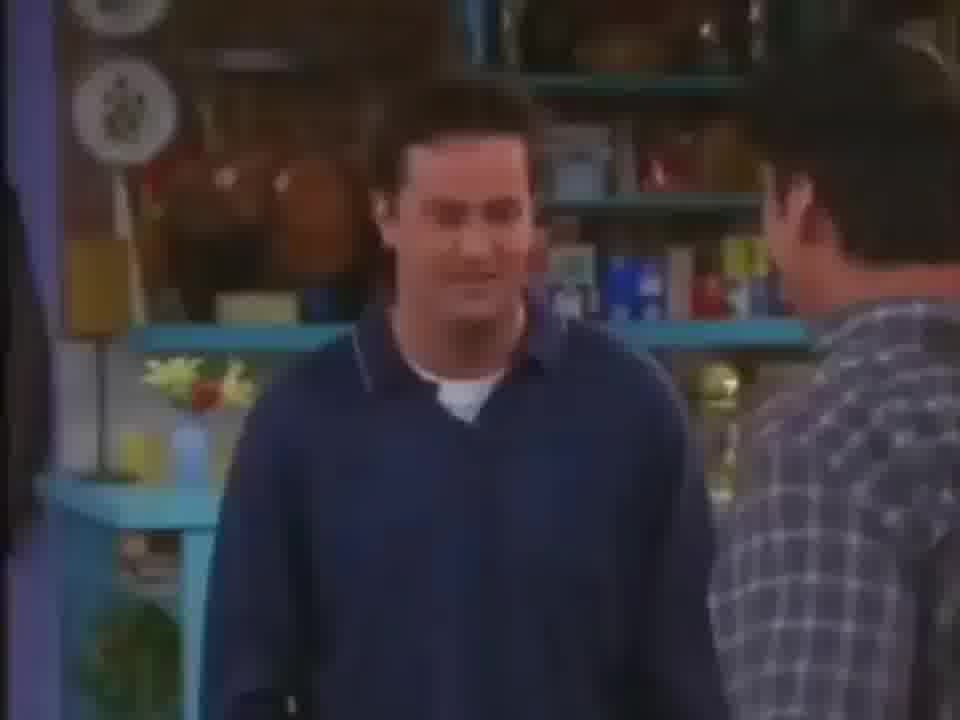}\vspace{1pt}
\includegraphics[width=0.15\textwidth]{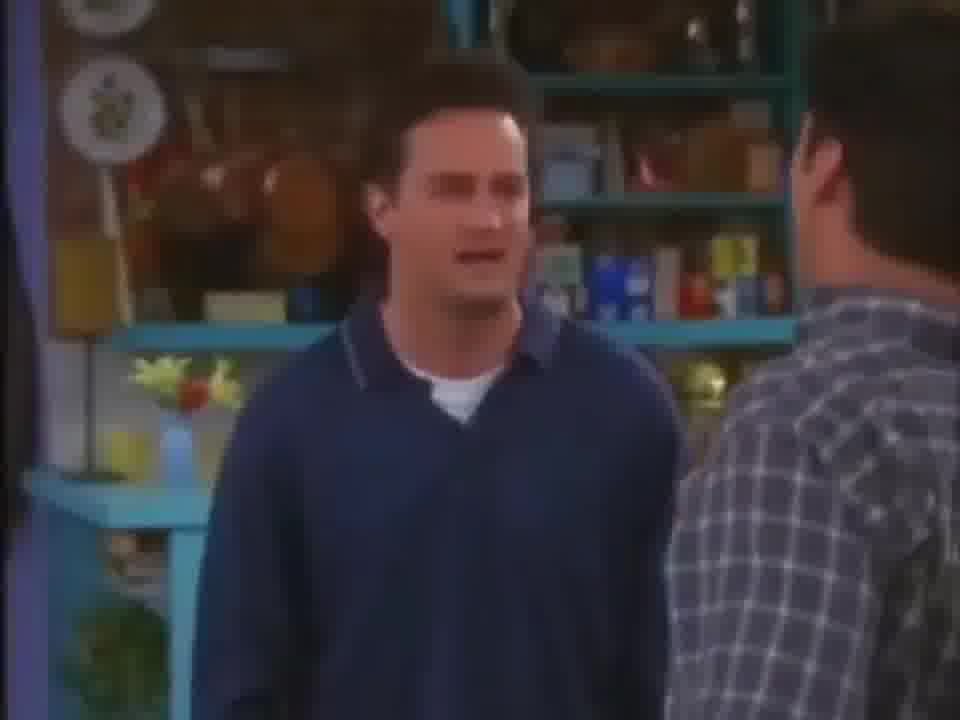}\vspace{1pt}
\includegraphics[width=0.15\textwidth]{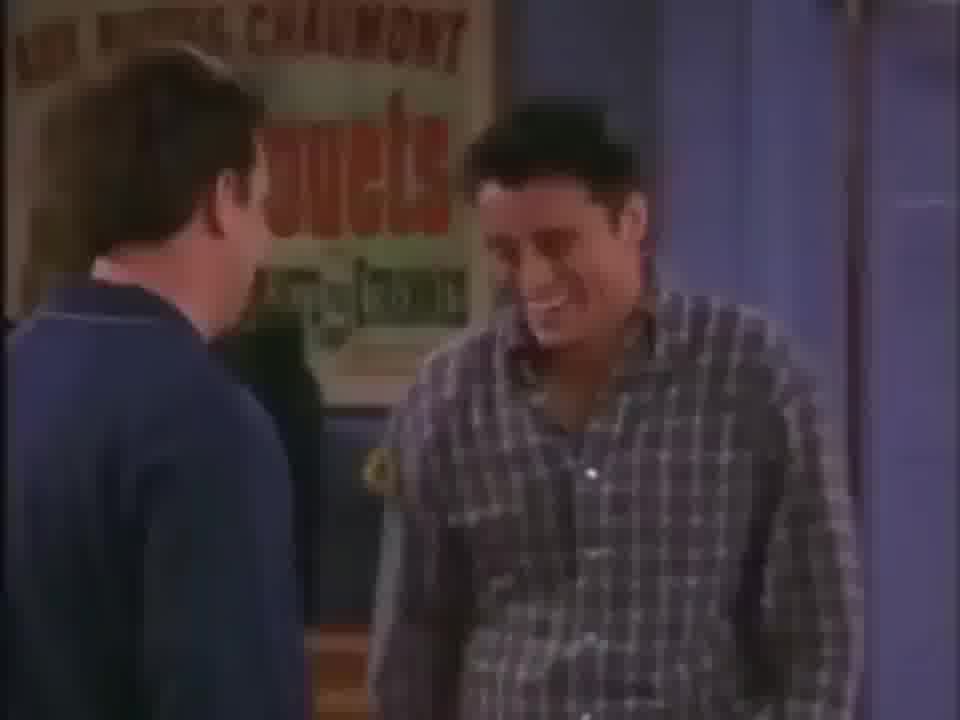}\vspace{1pt}
\end{minipage}

No.4
\begin{minipage}[c]{0.63\textwidth}
\includegraphics[width=0.15\textwidth]{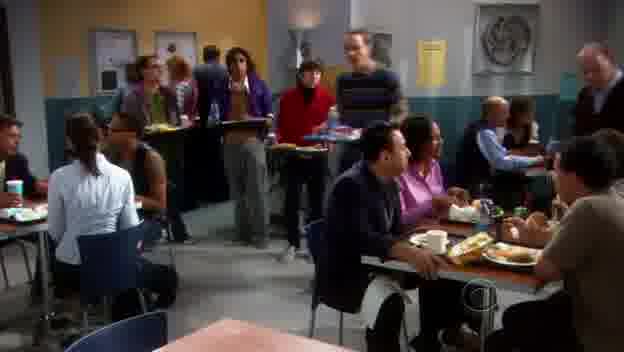}\vspace{1pt}
\includegraphics[width=0.15\textwidth]{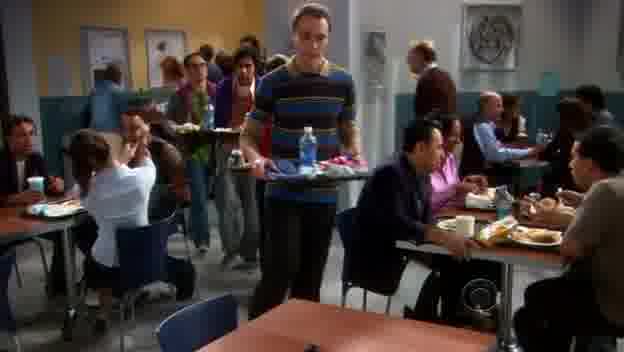}\vspace{1pt}
\includegraphics[width=0.15\textwidth]{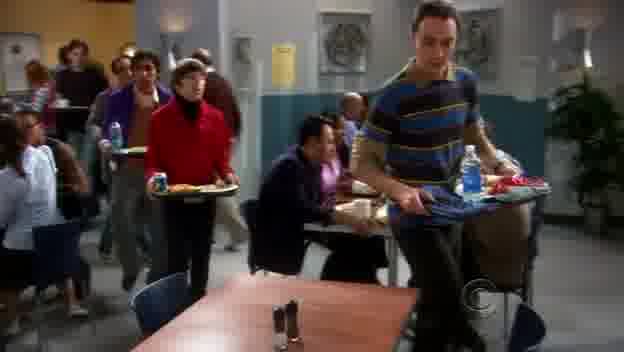}\vspace{1pt}
\includegraphics[width=0.15\textwidth]{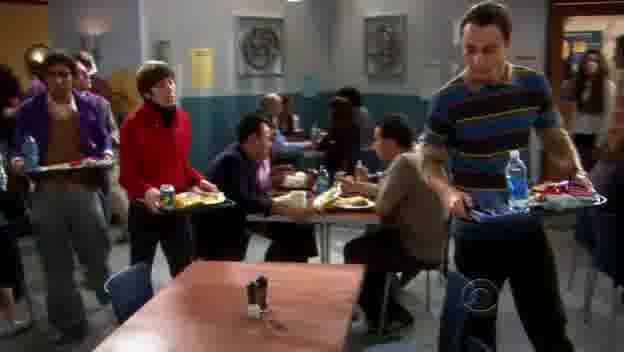}\vspace{1pt}
\end{minipage}

No.5
\begin{minipage}[c]{0.63\textwidth}
\includegraphics[width=0.15\textwidth]{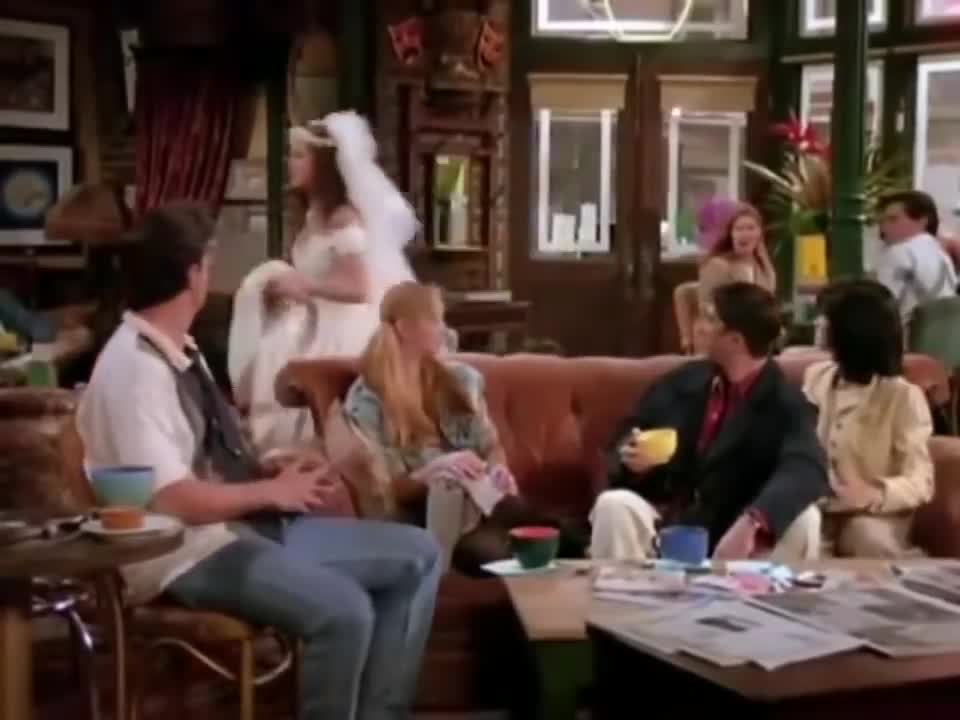}\vspace{1pt}
\includegraphics[width=0.15\textwidth]{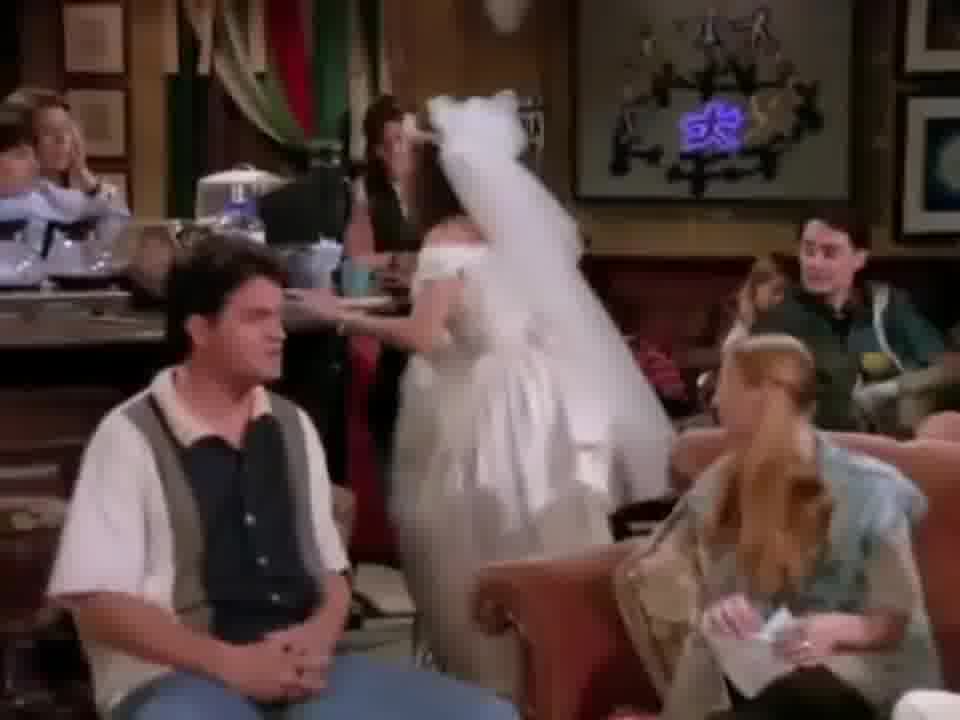}\vspace{1pt}
\includegraphics[width=0.15\textwidth]{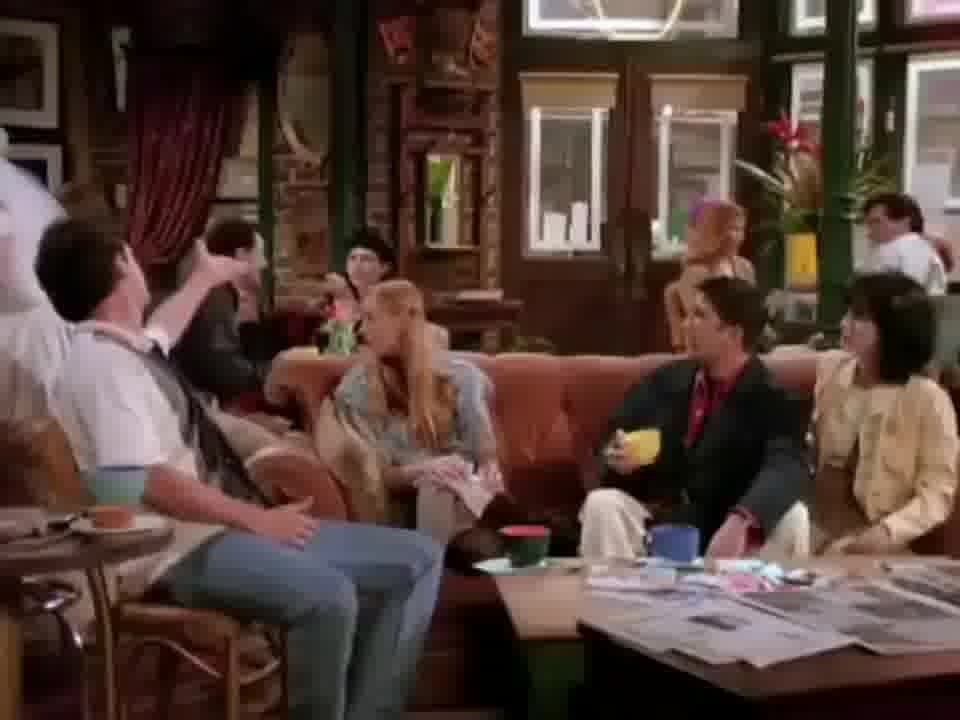}\vspace{1pt}
\includegraphics[width=0.15\textwidth]{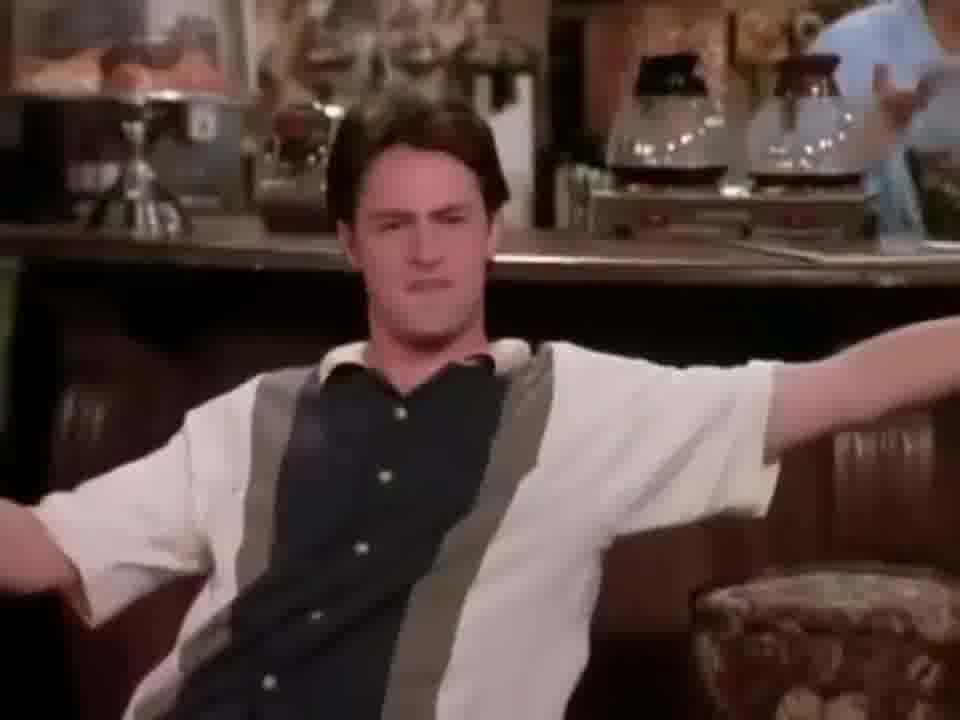}\vspace{1pt}
\end{minipage}
\caption{Misclassified utterances with corresponding video frames. Each line denotes an individual conversation corresponding the text in Table~\ref{tab:error}.}
\label{fig:cases}
\end{figure}

We perform an error analysis and show several typical mis-classification cases (textual utterance plus image), including the cases that MTL predicts correctly while STL fails, and that both setups fails to predict correctly. These cases are shown in Table~\ref{tab:error} and Fig.~\ref{fig:cases} .

From Table~\ref{tab:error} and Figure~\ref{fig:cases}, we found that mis-classification often happens in the situation where the speaker uses a few positive words to express his/her sarcastic attitude. In this case, QUIET first mistakenly treats it as a positive sentiment utterance, and thus feeds this wrong sentiment identification into the complex-valued embedding, then makes a wrong decision. Further, we also notice that few errors occur when an utterance expresses very negative sentiment. QUIET may mix up the negative sentiment or anger emotion with the sarcasm polarity. This is due to the subtle difference between sarcasm, sentiment, and emotion. Discriminating irony attitude from negative sentiment is a tricky and complex problem, which is still an open area of research. 

\begin{table*}[t]
\small
\begin{center}
\scalebox{0.85}{
\begin{tabular}{|c|c||c|c|c|}
\hline
\multirow{2}*{\textbf{No.}} & \multirow{2}*{\textbf{Utterances}} & \multicolumn{3}{c|}{\textbf{Sarcasm (T+V)}} \\ 
\cline{3-5}
& & Actual & STL & MTL \\\hline
\hline
1 & \textit{Good idea, sit with her. Hold her, comfort her. And if the moment feels right, see if you can cop a feel.} & S & NS & NS\\ 
\hline
2 & \textit{Just the latest copy of Applied Particle Physics quarterly.} & S & NS & NS\\
\hline
3 & \textit{Oh my god, you almost gave me a heart attack!} & S & NS & NS\\
\hline
4 &  \textit{I'm sorry, I am not going back to the Renaissance fair.} & NS & S & NS\\
\hline
5 & \textit{And I just won a million dollars!} & NS & S  & NS\\
\hline
\end{tabular}
}
\end{center}
\caption{\label{tab:error} Few error cases where MTL framework performs better than the STL framework.}
\end{table*}

\subsection{Discussion on Inter-Task Incompatibility}\label{sec5.8}
\begin{figure}[t]
{\centering
\scalebox{0.88}{
   \subfloat[]{\label{Fig:L1}
 \includegraphics[height=1.6in, width=1.5in]{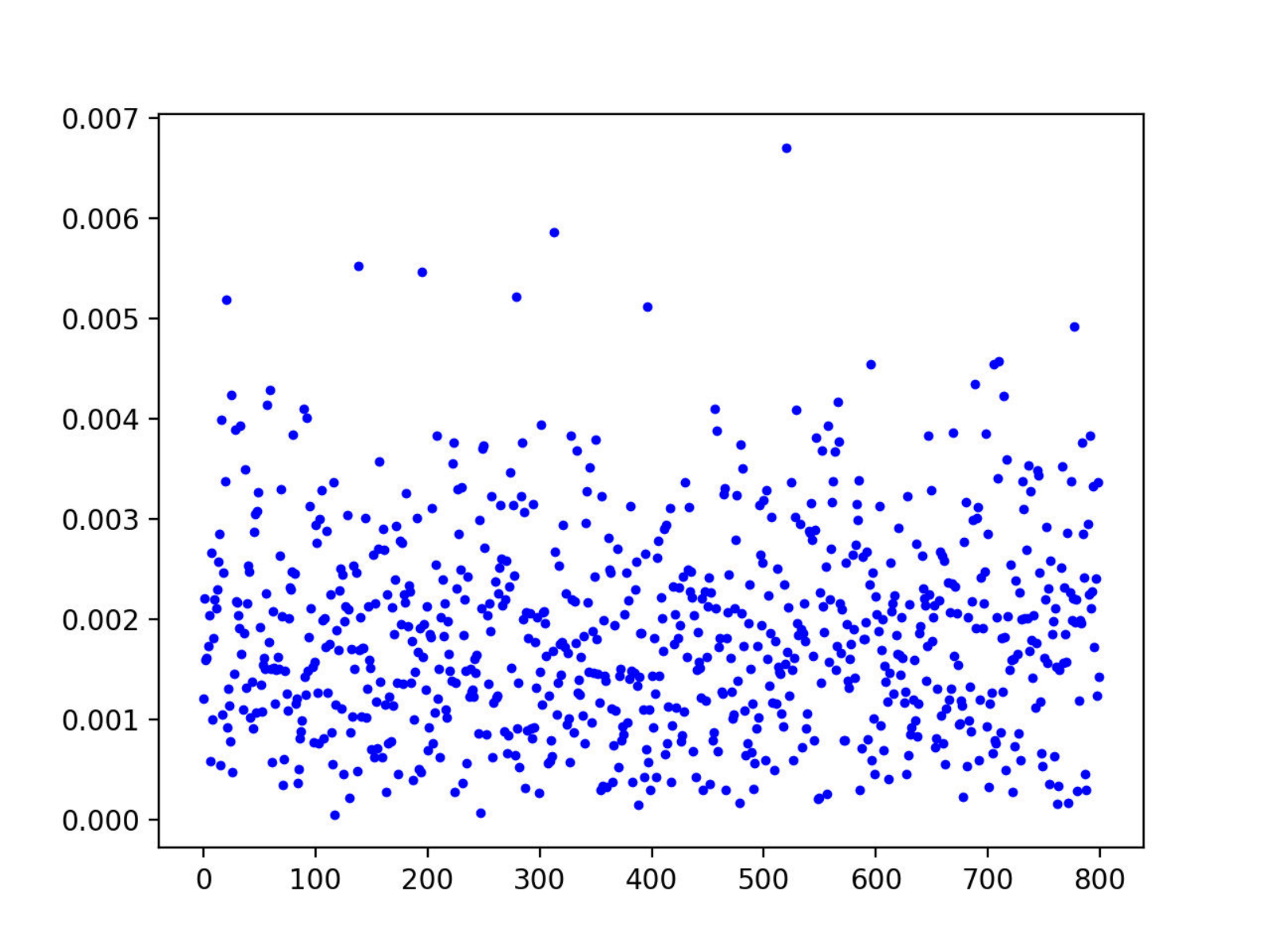}}~~~
  \subfloat[]{\label{Fig:R1}
  \includegraphics[height=1.6in, width=1.5in]{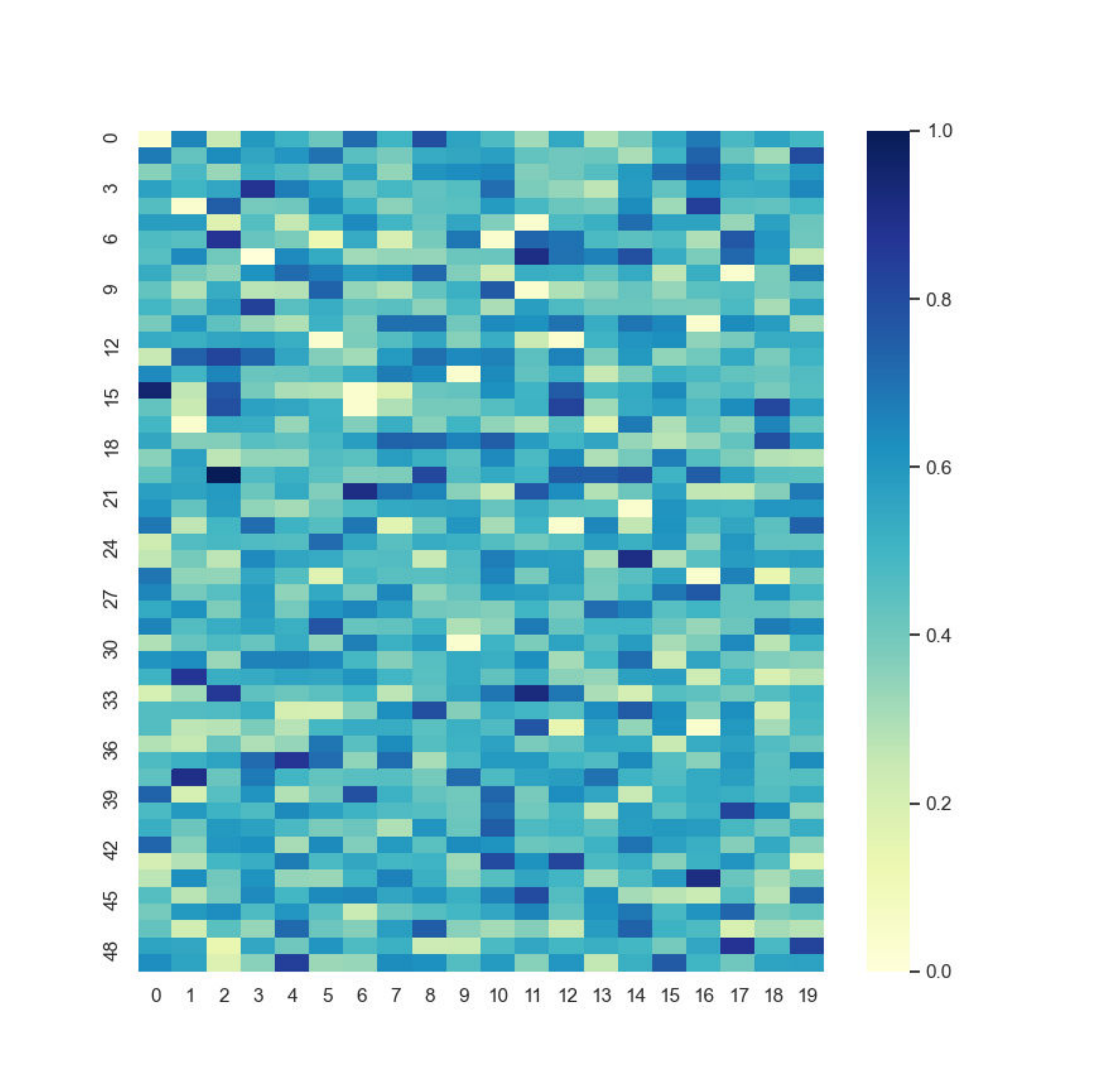}}
  }}

   \caption{Visualization of the commutation relation and quantum relative entropy.  }
\label{fig:incompatibility}
\end{figure}

\begin{table}[t]
\begin{center}
\scalebox{0.85}{
\begin{tabular}{|c|c|c|c|c|c|c|l|}
\hline
 \textbf{Dataset} & \textbf{Avg.}  & \multicolumn{6}{c|}{\textbf{Sample Correlation Scores}}    \\ \hline \hline
 MUStARD &  0.484 &   0.517 &  0.422 &  0.448 &  0.461 & 0.437  &  0.494 \\ \hline
\end{tabular}
}
\end{center}
\caption{\label{tab:corr} The correlation between sentiment, and sarcasm tasks.}
\end{table}

For a more detailed exploration of the incompatible measurement, we train 800 pairs of sentiment and sarcasm measurement operators, and calculate the commutation relation for each pair. The results are visualized in Figure~\ref{Fig:L1}. We can notice a violation of the commutation law, i.e., $\left [ M^{sar}_{\gamma},M^{sen}_{\delta } \right ]\neq0$ for all pairs, implying sentiment and sarcasm are incompatible. 
To further validate this observation, we introduce quantum relative entropy\footnote{$D(\sigma || \rho) = Tr\sigma log\sigma- Tr\sigma log \rho $. Here $ \sigma$ and $\rho $ are two measurement operators, $Tr$ means the trace operation.}, which is a kind of ``distance'' measure between quantum states, 
the smaller quantum relative entropy show the closer correlation between sentiment and sarcasm operators.
Average correlation and  sample correlation scores are presented in Table~\ref{tab:corr} and Figure~\ref{Fig:L1}, \ref{Fig:R1}, showing the two tasks are correlated. The result justifies the need of incompatible measurement and explains its effectiveness against traditional multi-task learning setting in Table~\ref{tab:ablation}.

Furthermore, an analysis of data shows that 84\% of sarcasm samples in MUStARD express explicit sentiments. In MUStARD, 38\% of ironic utterances are also positive. These results support our hypothesis that sarcasm and sentiment are closely related.

\begin{table*}[t]
\small
\begin{center}
\scalebox{0.85}{
\begin{tabular}{|c|c||c|c|c|}
\hline
\textbf{No.} & \textbf{Utterances} & \textbf{Sar.} & \textbf{Sent.} & \textbf{Emo.} \\\hline
\hline
a & \textit{Well, my beer isn't flat and my rack's not saggy. So far, the future's great.} & True & Pos. & Hp.\\ 
\hline
b & \textit{Boy, when you meet Bernadette, the field of robotics really took a hit.} & True & Pos. & Hp.\\
\hline
c & \textit{Wow, he really went where no man has gone before.} & False & Pos. & Sp.\\
\hline
d & \textit{No, but last year, at Magic Mountain, he got such a bad sunburn, we had to cut him out of it.} & False & Neg. & Sd.\\
\hline
\end{tabular}
}
\end{center}
\caption{\label{tab:casesstudy} Some cases that showing the interactions among sentiment, sarcasm, and emotion.}
\end{table*}

\begin{center}
\begin{figure}[t]
\scalebox{0.92}{
case a
\begin{minipage}[c]{0.69\textwidth}
\includegraphics[width=0.15\textwidth]{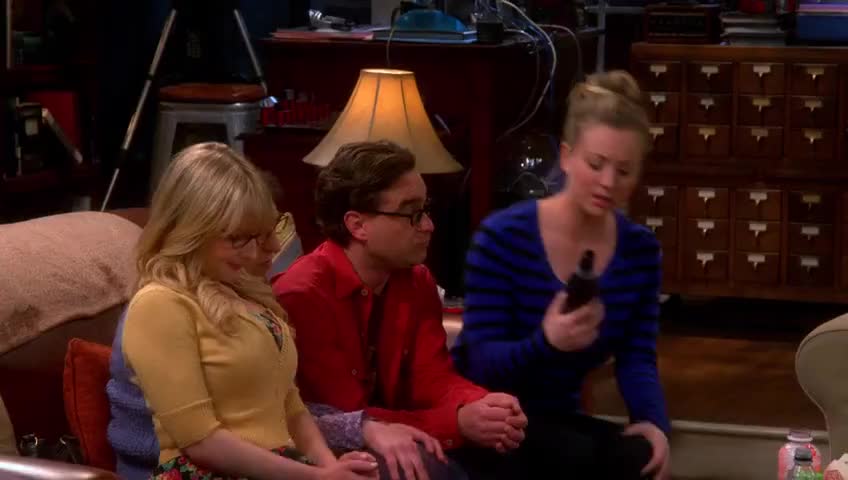}\vspace{1pt}
\includegraphics[width=0.15\textwidth]{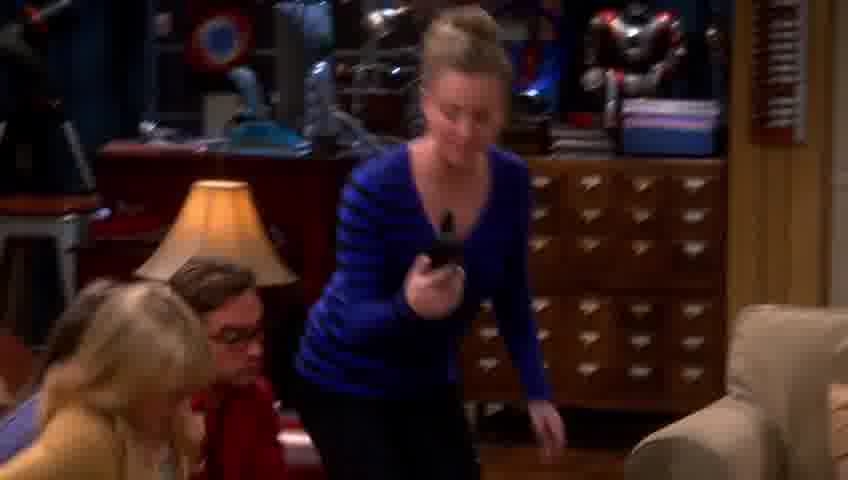}\vspace{1pt}
\includegraphics[width=0.15\textwidth]{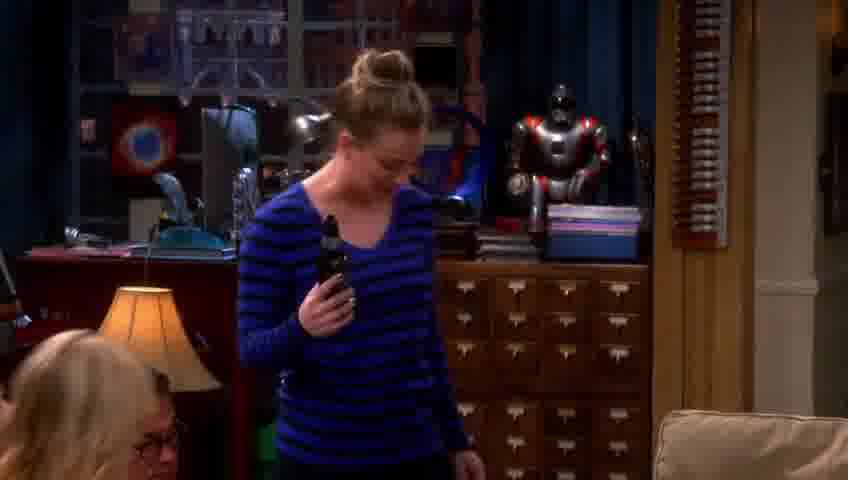}\vspace{1pt}
\includegraphics[width=0.15\textwidth]{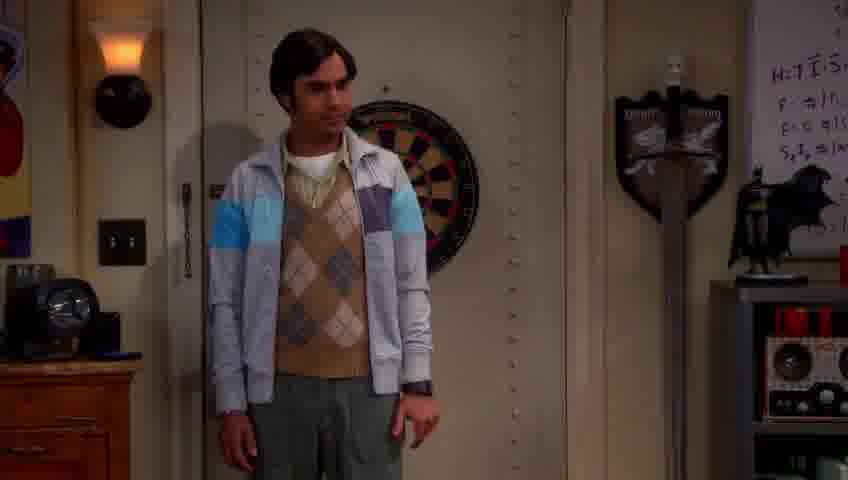}\vspace{1pt}
\end{minipage}

case b
\begin{minipage}[c]{0.69\textwidth}
\includegraphics[width=0.15\textwidth]{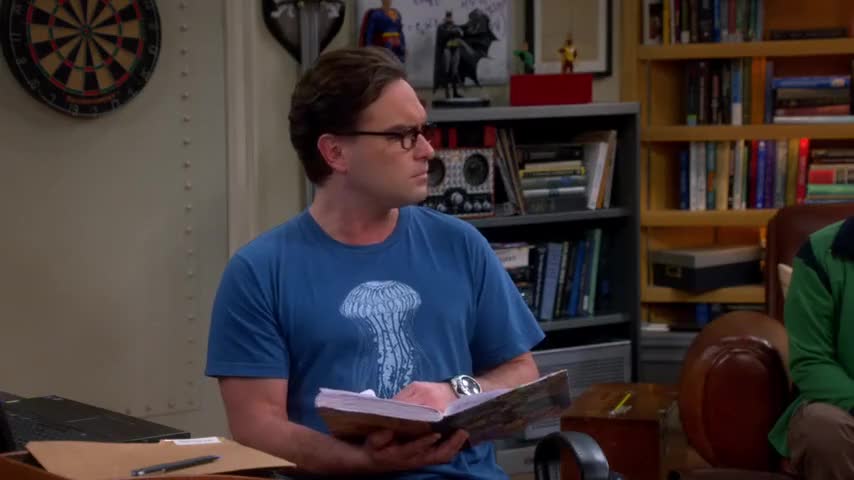}\vspace{1pt}
\includegraphics[width=0.15\textwidth]{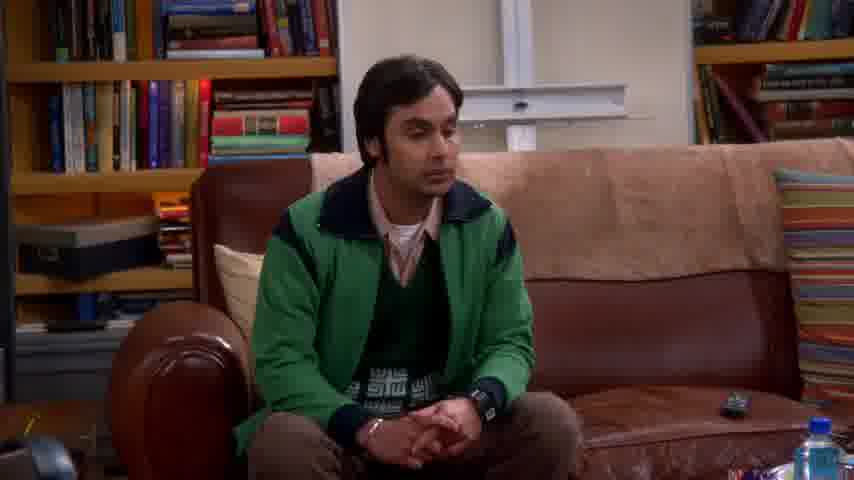}\vspace{1pt}
\includegraphics[width=0.15\textwidth]{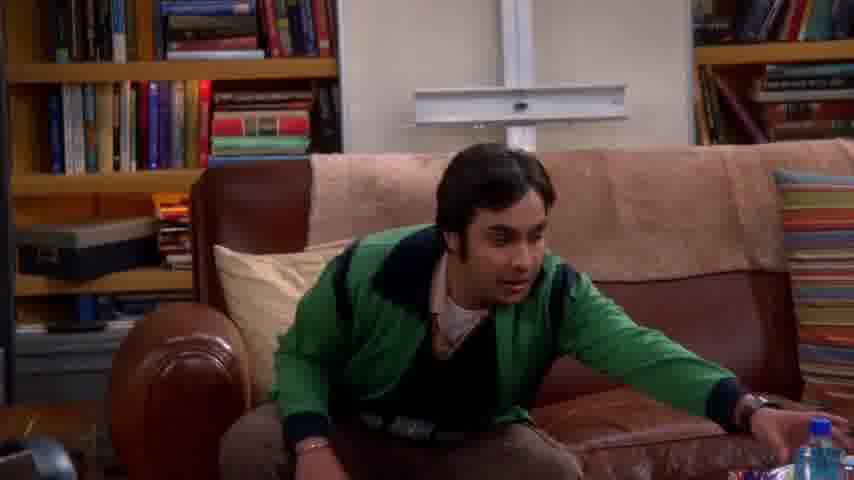}\vspace{1pt}
\includegraphics[width=0.15\textwidth]{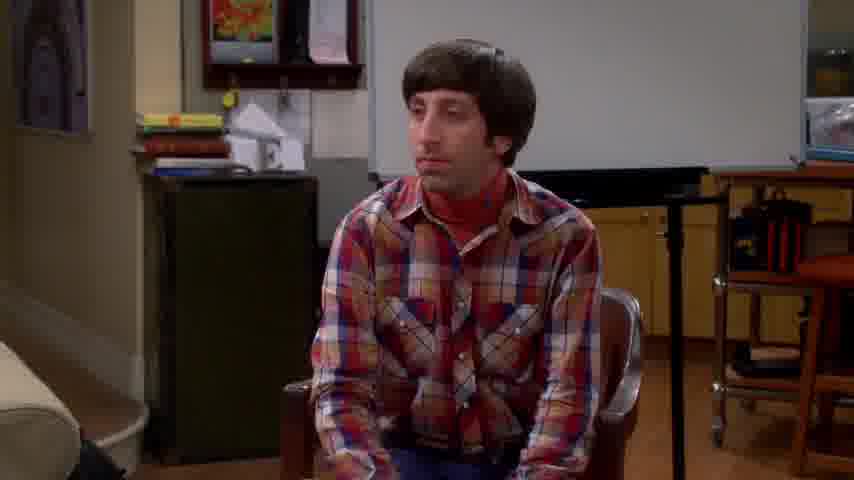}\vspace{1pt}
\end{minipage}

case c\,
\begin{minipage}[c]{0.69\textwidth}
\includegraphics[width=0.15\textwidth]{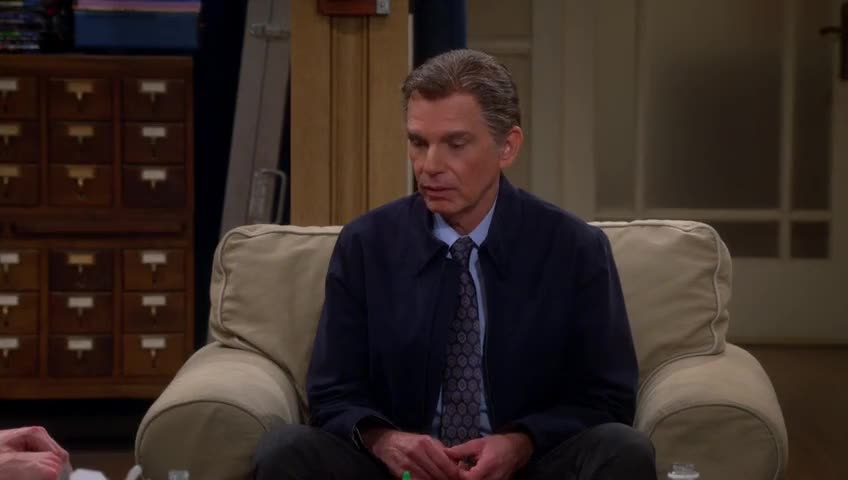}\vspace{1pt}
\includegraphics[width=0.15\textwidth]{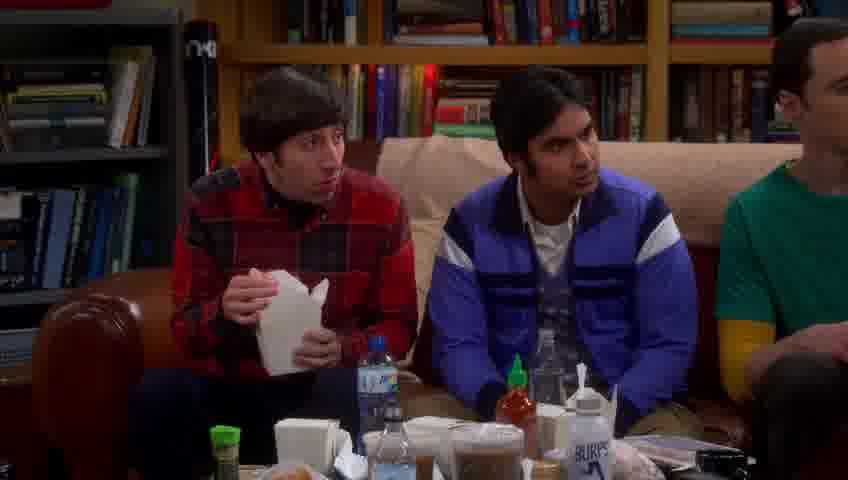}\vspace{1pt}
\includegraphics[width=0.15\textwidth]{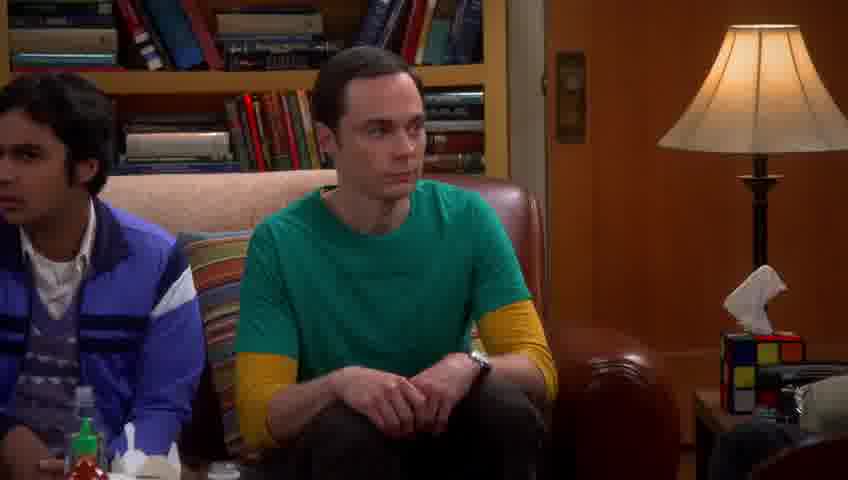}\vspace{1pt}
\includegraphics[width=0.15\textwidth]{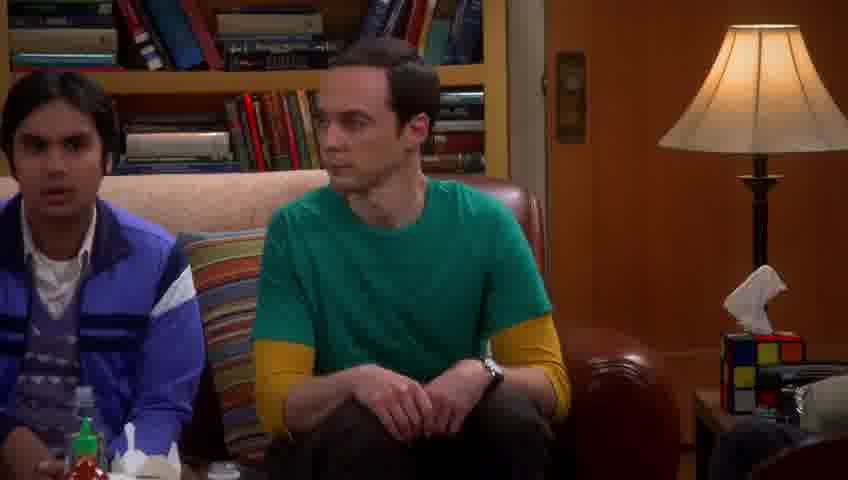}\vspace{1pt}
\end{minipage}

case d
\begin{minipage}[c]{0.69\textwidth}
\includegraphics[width=0.15\textwidth]{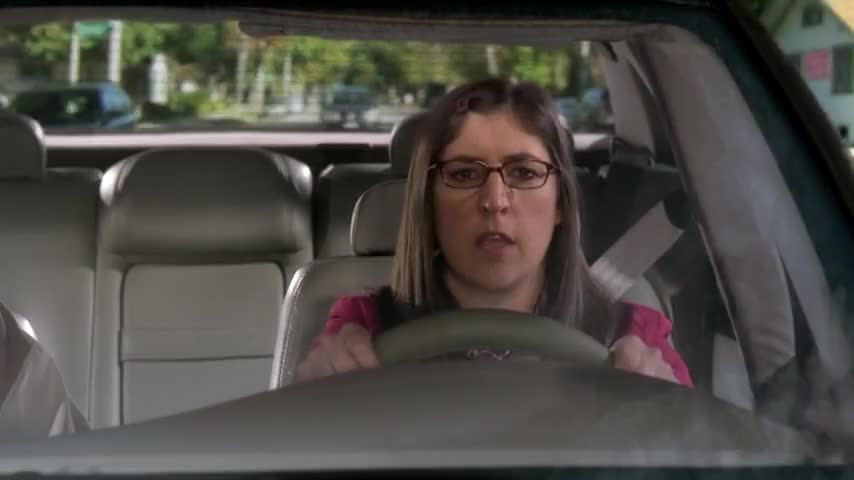}\vspace{1pt}
\includegraphics[width=0.15\textwidth]{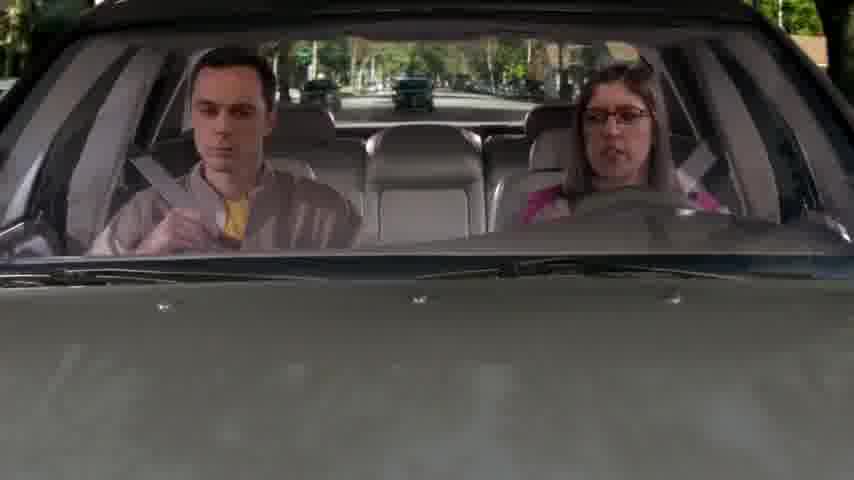}\vspace{1pt}
\includegraphics[width=0.15\textwidth]{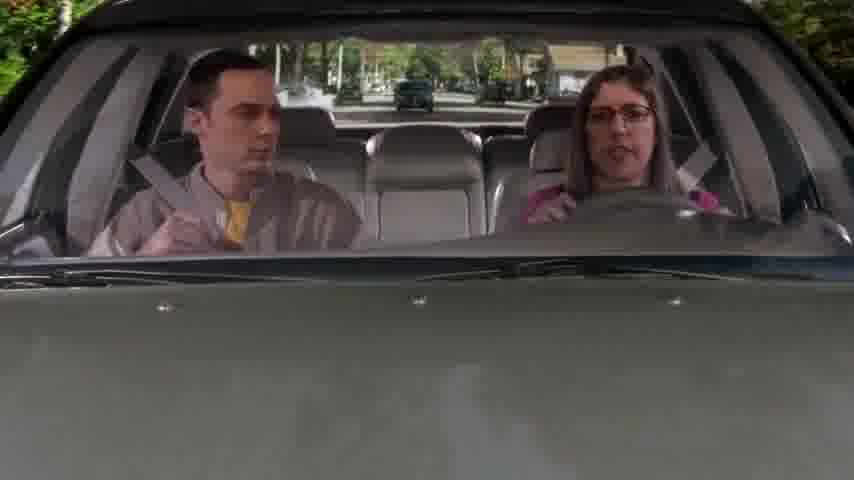}\vspace{1pt}
\includegraphics[width=0.15\textwidth]{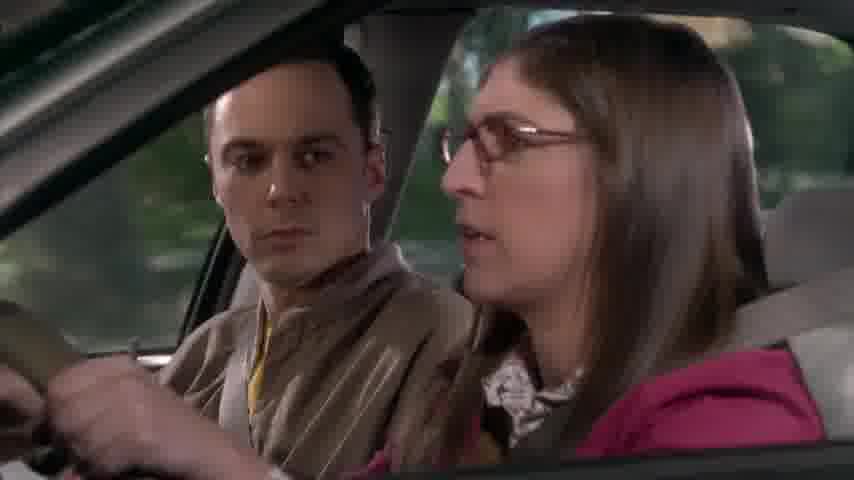}\vspace{1pt}
\end{minipage}}
\caption{Figures for cases in Table~\ref{tab:casesstudy}}
\label{fig:casesstudy}
\end{figure}
\end{center}

\subsection{Case Study}
We present a few classical cases in Table~\ref{tab:casesstudy}. We can notice that sarcasm detection reaps the greatest benefit from the other two tasks. Through incorporating the sentiment and emotion information, QUIET often makes correct decision on sarcasm detection. One possible reason is that sarcasm involves a higher level of abstraction and more subjectivity. By comparing (a)(b) and (c)(d), we see that  sentiment analysis offers the greatest help to emotion recognition while emotion recognition may benefits sarcasm detection more. The reason is that the facial expression and the gesture may help detect sarcasm. In contrast, emotion also helps sentiment. But sarcasm detection plays the least role in understanding emotion. Hence, it may be a reasonable choice to place sarcasm
detection as the main task, as we did.

\begin{table}[t]
\begin{center}
\scalebox{0.86}{
\begin{tabular}{|c|c|c|c|}
\hline
 \textbf{Model} & \textbf{Sar.} & \textbf{Sent.} & \textbf{Emo.}    \\ \hline \hline
 triple-slit model  &  72.43 &   62.71 &  32.44  \\ \hline
 double-slit model  &  83.70 &   74.89 &  37.69  \\ 
 $\bigtriangleup$ & (+11.27\%)  & (+12.18\%) &  (+5.25\%) \\ \hline
\end{tabular}
}
\end{center}
\caption{\label{tab:double-triple} Comparison of results on double-slit model and triple-slit model. Compared by micro-F1 scores.}
\end{table}

\subsection{Comparison with Previous Works}
We argue that our proposed model is quite different from all of the previous QP based models (including our previous works)~\cite{liuetal2021smilemean,9400728}. In this work we make the first attempt to introduce three modalities (textual, visual and acoustic modality) under a unified QP driven framework. The experiment results in Table~\ref{tab-sml}, Table~\ref{tab-bml} and Table~\ref{tab:tml} show the improvement by introducing new modality in to the framework. What's more, under the inspiration from work \cite{liuetal2021smilemean} that quantum incompatible measurement can handle the interaction between different tasks and under the multi-task learning framework, one task can benefit others. Thus we introduce the emotion recognition as the third task. In the quantum composition layer, we use GRU to learn local contextual interaction which are then encapsulate into a density matrix to represent both long and short contexts. 

We elaborately design the quantum multi-modal fusion layer. In this framework, we have information from three modalities. However, the double-slit experiment only involves two propagation paths. So we propose three single quantum interference-like fusion component for three different combinations of modalities ($tv$,$ta$,$va$). 


In addition, we also test the triple-slit interference component. However, the triple-slit model performs not very well, and does not outperform the double-slit model, where the experimental results are 0.72, 0.62 and 0.32 on micro-F1 for sarcasm detection, sentiment analysis and emotion recognition tasks respectively. However, results on double-slit model are 0.83, 0.74, and 0.37, double-slit model overcome triple-slit model on all three tasks, result are shown in Table~\ref{tab:double-triple}. Hence, the triple-slit inspired model is not as good as double-slit inspired model. The reason is that simply extend the quantum interference fusion approach to triple-slit scenario will introduce more noisy information. The interaction across three modalities is more complex. In addition, the triple-slit interference experiment does not exist in quantum physics, where such attempts would compromise the theoretical interpretability of our model. Hence, we choose to keep the current bi-modal fusion approach due to the above-mentioned reasons. 

After multi-modal fusion, the quantum imcompatible measurement layer is used to measure three tasks simultaneously. Modeling the correlation across three tasks is more difficult than modeling the bi-task correlation. The innovation lies on how to design the number of measurement operators and how to extend the commutation relation to measure the correlation across three tasks.

Additionally, we conduct detailed experiments. We list the result on all three tasks and study the influence caused by different task combinations, component combinations, context ranges and modalities. We also make the error analysis and case study. The experimental results can strongly prove the reliability of the theory and the effectiveness of the proposed model.

\section{Conclusions and Future Work}
Joint multi-modal sarcasm, sentiment, and emotion analysis is a relatively unexplored task in NLP and affective computing. Inspired by the recent success of QP in modeling human cognition and decision making, we take the first step to introduce QP into the task. We thus propose a quantum probability driven framework for multi-modal sarcasm, sentiment, and emotion analysis, namely QUIET. It consists of a complex-valued multi-modal encoder, a quantum composition layer, a quantum interference-like inter-modal fusion layer and a quantum measurement layer. The main idea is to represent each multi-modal utterance in a conversation as a complex-valued vector and then perform multi-modal fusion via quantum interference. Finally, quantum incompatible measurements are performed on the multi-modal representation to yield the probabilistic outcomes of sarcasm, sentiment, and emotion recognition. We empirically prove the effectiveness of the proposed model by outperforming the state-of-the-art baselines. 

Given the limited availability of multi-modal datasets providing labels for sarcasm, sentiment, and emotion at the same time, we evaluated the proposed model on a benchmark dataset that is the only one currently satisfying the above requirement. The effectiveness of QUIET needs to be further tapped. To this end, we plan to create a larger scale multi-modal multi-task conversational affect dataset to advance the development of multi-modal sarcasm, sentiment, and emotion joint analysis. 
Moreover, the GRU-based structure used in the proposed model takes sequential data as input. It can only calculates from left to right or from right to left, limiting the parallel computing ability of the model. To alleviate the problem, we plan to investigate a quantum inspired transformer structure to better capture the correlations among utterances and improve the model's parallel computing ability.

\ifCLASSOPTIONcaptionsoff
  \newpage
\fi




\bibliographystyle{IEEEtran}
\bibliography{IEEEabrv, custom}
%



%

\begin{IEEEbiography}[{\includegraphics[width=1in,height=1.25in,clip,keepaspectratio]{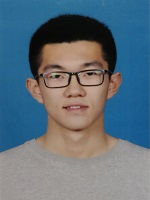}}]{Yaochen Liu}
is a graduate student in the School of Computer Science, Beijing Institute of Technology in China. He received his bachelor's degree from Northeastern University, Shenyang. His current research direction is quantum probability theory driven sarcasm detection model. Besides using the QP, he's also interested in multi-modal framework.
\end{IEEEbiography}

\begin{IEEEbiography}
[{\includegraphics[width=1in,height=1.25in,clip,keepaspectratio]{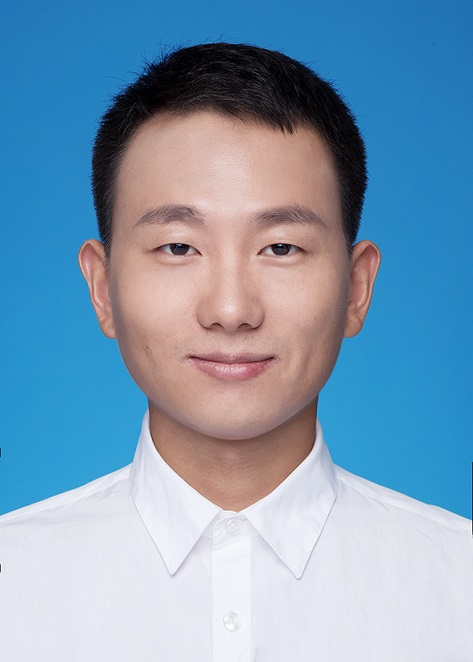}}]
{Yazhou Zhang}
received his Ph.D. degree in the College of Intelligence and Computing from Tianjin University (Tianjin, China) in 2020. He is currently a Lecturer in Software Engineering College at Zhengzhou University of Light Industry (Zhengzhou, China), and also a Postdoctoral Fellow in the School of Nursing at The Hong Kong Polytechnic University. He was a Postdoctoral Fellow in Artificial Intelligence Laboratory at Tianjin University-China Mobile Communication Group Tianjin Co., Ltd. in 2022. His research interests include opinion mining (or sentiment analysis), data fusion, and quantum cognition. He is currently working on developing quantum inspired sentiment analysis models and their application to problems like conversational sentiment analysis, information fusion and evolution of user emotional state.
\end{IEEEbiography}

\begin{IEEEbiography}
[{\includegraphics[width=1in,height=1.25in,clip,keepaspectratio]{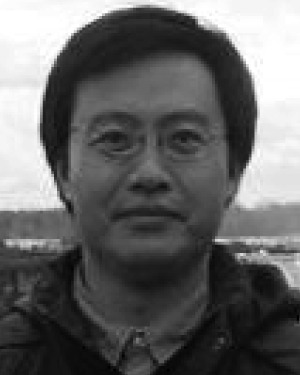}}]
{Dawei Song} received his PhD (Information Systems) from the Chinese University of Hong Kong in 2000. He is currently a professor at Beijing Institute of Technology. Prior to this appointment, he was a professor at Tianjin University (2012-2018), and a Professor of Computing at the Robert Gordon University, UK (2008-2012), where he remains as an Honorary Professor since 2012.
He has also worked as a Senior Lecturer at the Knowledge Media Institute of The Open University, UK (2005-2008), where he remains as a part-time professor since 2012; and as a Research Scientist (since 2000) and Senior Research Scientist (since 2002) at the Cooperative Research Centre in Enterprise Distributed Systems Technology, Australia. 
His research interests include theory and formal models for natural language and multi-modal information processing, and user-centric information seeking.
\end{IEEEbiography}







\end{document}